\RequirePackage[2020-02-02]{latexrelease}
\documentclass{clv3}

\usepackage{comment}

\let\oldnumdef\numdef
\let\numdef\relax
\usepackage{hyperref}
\renewcommand{\numdef}{\oldnumdef}

\usepackage{CJKutf8} 

\usepackage{subcaption}
\usepackage[dvipsnames]{xcolor} 
\usepackage{booktabs} 
\usepackage{linguex}
\definecolor{darkblue}{rgb}{0, 0, 0.5}
\hypersetup{colorlinks=true,citecolor=darkblue, linkcolor=darkblue, urlcolor=darkblue}
\usepackage{float}
\usepackage{enumitem}
\usepackage{tikz-dependency}
\usepackage{amsmath}
\usepackage[euler]{textgreek}
\usepackage{color, colortbl}
\usepackage[poorman]{cleveref}

\definecolor{LightCyan}{rgb}{0.88,1,1}
\definecolor{LightGreen}{rgb}{0.76,1,0.72}
\definecolor{LightOrange}{rgb}{1,0.91,0.65}
\definecolor{LightPurple}{rgb}{0.627, 0.125, 0.941}
\definecolor{LightYellow}{rgb}{1,0.98,0.74}
\definecolor{LightPink}{rgb}{0.98,0.74,1}
\definecolor{LightGrey}{rgb}{0.72,0.82,0.87}
\definecolor{LightTan}{rgb}{0.83,0.80,0.56}

\newcommand{\midsepremove}{\aboverulesep = 0mm \belowrulesep = 0mm} \newcommand{\midsepdefault}{\aboverulesep = 0.605mm \belowrulesep = 0.984mm}

\begin{document}

\issue{50}{3}{2024}


\runningtitle{eRST: A Signaled Graph Theory}

\runningauthor{Zeldes et al.}

\pageonefooter{Action editor: Min Zhang. Submission received: 22 November 2023; revised version received: 18 March 2024; accepted for publication: 19 July 2024.}

\newcommand{\theory}{eRST}

\title{\theory{}: A Signaled Graph Theory of Discourse Relations and Organization}

\author{Amir Zeldes\thanks{1421 37th St. NW, Washington, DC 20057, USA. E-mail: amir.zeldes@georgetown.edu.}}
\affil{Georgetown University}

\author{Tatsuya Aoyama}
\affil{Georgetown University}
    
\author{Yang Janet Liu}
\affil{Georgetown University}

\author{Siyao Peng\thanks{Work done partly at Georgetown University.}}
\affil{LMU Munich}

\author{Debopam Das}
\affil{\r{A}bo Akademi University}

\author{Luke Gessler$^{\ast{}\ast{}}$}
\affil{University of Colorado Boulder}

\maketitle

\vspace{-5pt}

\begin{abstract}
In this article we present Enhanced Rhetorical Structure Theory (\theory{}), a new theoretical framework for computational discourse analysis, based on an expansion of Rhetorical Structure Theory (RST). The framework encompasses discourse relation graphs with tree-breaking, non-projective and concurrent relations, as well as implicit and explicit signals which give explainable rationales to our analyses. We survey shortcomings of RST and other existing frameworks, such as Segmented Discourse Representation Theory (SDRT), the Penn Discourse Treebank (PDTB) and Discourse Dependencies, and address these using constructs in the proposed theory. We provide annotation, search and visualization tools for data, and present and evaluate a freely available corpus of English annotated according to our framework, encompassing 12 spoken and written genres with over 200K tokens. Finally, we discuss automatic parsing, evaluation metrics and applications for data in our framework.
\end{abstract}

\section{Introduction}\label{sec:intro}

Natural language documents are more than just an ordered list of equally important and self-contained sentences: They form complex structures that can often be divided into more or less prominent sections and subsections, which together give rise to meanings that are not necessarily localizable to individual propositions by themselves. Identifying these structures and the meanings associated with them is the task of \textit{Discourse Parsing}, in which arbitrary documents are assigned an analysis within a theoretical parsing framework that defines the types of combinatory semantic and pragmatic meanings to be recognized, and the structures that components of a document can create.

While not discussed in the field of NLP as often as syntactic parsing or entity recognition, discourse parsing has been one of the `textbook' examples of Natural Language Understanding \cite[536--540]{JurafskyMartin2024} for a long time, with implementable frameworks being suggested as early as \namecite{MannThompson1988}'s Rhetorical Structure Theory (RST). Most approaches to discourse parsing involve (at least) recognizing  spans of text which are connected by one of a set of predetermined discourse relation types \cite{hovy-1990-parsimonious}, such as \textsc{cause} or \textsc{concession}, and naming the relation and configuration in which those parts appear in the text, which can take on many linguistic forms. For instance, in example \labelcref{ex:fell} from \citet[136]{AsherLascarides2003}, both formulations in a. and b. are typically interpreted to mean that the predicate \textit{pushed} is the cause of the predicate \textit{fell}, and that the pushing preceded the falling in time, although these events are related in chronological order in b., but in counter-chronological order in a.

\ex.\label{ex:fell} \a. Max fell. John pushed him.
  \b. John pushed Max. He fell.

The exact nature and inventory of such relations, sometimes called `coherence relations', `prominence relations', or also `rhetorical relations', as well as the structures they form, vary across theoretical accounts. 

Like other areas of NLP, discourse parsing has benefited from increasingly accurate scores following the introduction of large pre-trained language models, with scores approaching human performance on some subtasks, such as discourse unit segmentation \cite{GesslerEtAl2021}, recognition of explicitly signaled relations \cite{knaebel-2021-discopy}, as well as hierarchical parsing, especially for English in the news domain \cite{guz-etal-2020-unleashing,liu-etal-2021-dmrst,kobayashi-etal-2022-simple}. 

By contrast, less progress has been made in advancing our theories of discourse relations and their organization. After the introduction of RST and subsequent projects to construct datasets using the theory \citep{CarlsonEtAl2001}, several alternative frameworks were proposed to address some of its shortcomings (surveyed below in Section \labelcref{subsec:rst}), with the main strands resulting in implemented datasets including 
Segmented Discourse Representation Theory (SDRT, \citealt{AsherLascarides2003}, Section \labelcref{subsec:sdrt}),
the Penn Discourse Treebank framework (PDTB, \citealt{PrasadEtAl2006}, Section \labelcref{subsec:pdtb}),
and the Cognitive Approach to Coherence Relations (CCR, \citealt{SandersSpoorenNoordman1992}, Section \labelcref{subsec:ccr}).
These frameworks each improve on certain problems identified quite early on in RST, including notably:

\begin{itemize}
   \item Tree-breaking, non-projective structures (SDRT)
    \item Distinguishing implicitly and explicitly signaled relations, with the latter being more reliably identifiable (PDTB)
   \item Support for multiple concurrent relations (mainly SDRT, but to some extent all of the above)
    \item Identification of relation hierarchies or subtypes based on formal markers (PDTB) or feature structures (CCR)
    \item Explicit support for nested relations (SDRT)
\end{itemize}

Although there has been substantial work in each framework, including refinements to guidelines or covered phenomena, and development of new annotated resources, little has changed in the landscape of implemented theoretical models of discourse relations since the inception of PDTB almost two decades ago. However this stability should not be taken as a sign that our theoretical models are now completely satisfactory: Each of the theories mentioned above has shortcomings, such as inability to model hierarchical structure in PDTB, or lack of relative prominence marking in SDRT. 

In this article we aim to push the development of discourse representation theories further, by proposing a new formalism that draws on insights from several frameworks in an attempt to keep the most useful parts of the original formulation of discourse parsing as envisioned by \citet{MannThompson1988}, while incorporating solutions to problems from over three decades of work in the field. Since our formalism is `backwards compatible' with RST, we designate it \textit{Enhanced Rhetorical Structure Theory} (\theory{}), in the hopes of drawing researchers already familiar with RST and harnessing existing resources for its development (in this sense it can be viewed as an optional `enhanced' representation, similar to Enhanced Dependencies for more basic Universal Dependencies in syntax, \citealt{nivre-etal-2020-universal}). At the same time, our framework offers important additional expressive mechanisms that should appeal to researchers engaged with other frameworks, specifically supporting:

\begin{itemize}
    \item Multiple relations between the same nodes
    \item Non-projective, tree-breaking structures
    \item Maintaining RST's recursive prominence hierarchy despite the above
    \item Marking categorized and subtyped discourse relation signals, including implicit and explicit connectives, as well as alternative lexicalization mechanisms
    \item Use of a hierarchical relation taxonomy
    \item Supporting new NLU applications by linking relations to implicated spans of text fulfilling specific relation participant roles
\end{itemize}

\noindent We would like to stress that while the last point is of interest to us, the primary motivation for \theory{} is not improving performance on any particular NLP tasks compared to RST, but simply to provide a more comprehensive and detailed representation of discourse relations in text across any genre, which can recover relations that are present, but not currently covered by RST analyses, along with the rationale or evidence supporting and sub-categorizing their occurrences.

\theory{}, with its advanced set of features, can support the inquiry of numerous discourse phenomena. Some general research questions we envisage \theory{} would help us investigate include how do discourse relations and their signals are distributed across texts and text types or genres? What correlations exist between relation and signal types? Are there semantic or pragmatic correlates of the amount and type of signaling observed for relation types? When and how often does natural discourse violate strict tree constraints? And to what extent can discourse relation identification be completely motivated by localizable signals? Complementing the theoretical framework proposed in this article, we also release data and tools to support development work, which are meant to cover a wide range of scenarios and types of text. The main contributions of this article are therefore:

\begin{enumerate}
    \item A new framework for discourse relation and discourse structure annotation
    \item Extending a robust freely available annotation tool to create gold standard data
    \item A corpus of over $200$K tokens covering $12$ spoken and written English text types\footnote{Since submission of this paper, the corpus has grown even larger and now covers $16$ genres, supplemented by test data in $8$ additional genres, for a total of $246$K tokens in $24$ genres; see Section \labelcref{sec:data} for more details.}
    \item A corresponding XML format, annotation and conversion tools for the freely available search and visualization tool ANNIS \cite{KrauseEtAl2016}
    \item A newly defined discourse parsing task including metrics and an official scorer
    \item A baseline system using a contemporary neural architecture and scores
\end{enumerate}

\section{Related Work}\label{sec:related-work}

\begin{figure}
    \centering
    \includegraphics[width=0.45\textwidth]{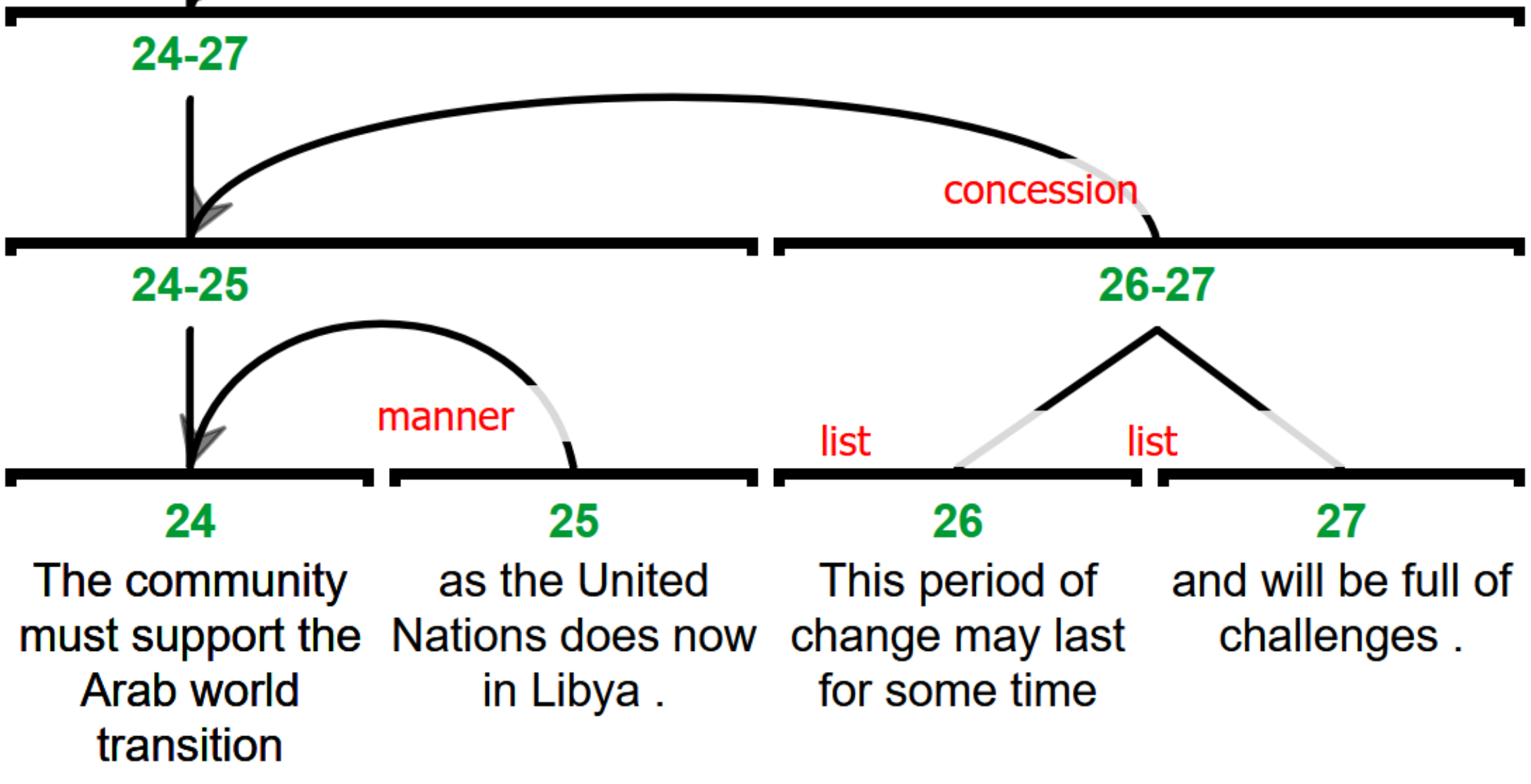}
    \caption{RST Fragment from GUM \cite{Zeldes2017}. The most central point is the nucleus in [24], to which other units are direct or indirect satellites (\textsc{manner} and \textsc{concession}). Symmetrical relations such as \textsc{list} are multinuclear nodes ([26]-[27]).}
    \vspace{-22pt}
    \label{fig:gum-rst-example}
\end{figure}

\namecite{MannThompson1988}'s formulation of discourse parsing understood relations to hold recursively between adjacent and contiguous spans of text, which covered entire arbitrary documents down to the level of basic propositions known as Elementary Discourse Units (EDUs), thereby forming a hierarchical labeled tree, as in Figure \labelcref{fig:gum-rst-example}. Labeled RST trees are directed and assume a distinction between more prominent `nucleus' units, and less prominent `satellite' units at each level of the tree.

The recursive nature of RST trees was particularly appealing to early research on automatic summarization \cite{Marcu1997,TeufelMoens2002} and dialog planning \cite{moore-paris-1993-planning,TaboadaLavid2003}, since removing satellites and their descendants could be used to obtain extractive summaries of arbitrary passages \cite{liu2024generalizability} (e.g.~[24] is the best extractive summary unit for the entire tree in Figure \labelcref{fig:gum-rst-example}) and a recursive tree could be used to track complex bifurcating topics in a long conversation.

In the years since the proposal of RST, a number of competing frameworks, which will be surveyed below in more detail, have suggested both limiting and expanding the scope of discourse relation identification. For example, according to \citet[2]{SandersSpoorenNoordman1992}'s Cognitive approach to Coherence Relations (CCR), relations should be identified by the presence of meanings ``of two or more discourse segments that cannot be described in terms of the meaning of the segments in isolation'', without necessarily assuming a hierarchy or coverage of the text, and are distinguished using a set of binary attributes (e.g.~basic vs. non-basic ordering in example \labelcref{ex:fell} above, see \citealt{HoekEvers-VermeulSanders2019}). \citeauthor{AsherLascarides2003}'s Segmented Discourse Representation Theory (SDRT), proposed that segments could participate in multiple relations, addressing early criticism of RST's strict tree constraint \cite{moore-pollack-1992-problem}, and forming a graph rather than a tree, with elementary units which are also allowed to nest. SDRT distinguishes subordinating and coordinating relations, rather than distinguishing satellites from nuclei, with some consequences for the structures postulated by the theory.

Moving in the opposite direction and more similarly to CCR, the framework of the Penn Discourse Treebank (PDTB, \citealt{PrasadEtAl2006}) proposed to identify relations as projections of explicit or implicit discourse markers called \textit{connectives}, such as the word `because', whose presence (or possible presence when omitted) indicates a causal relation. PDTB analyses are also called shallow discourse parses \cite{xue-etal-2016-conll}, since they do not assume a hierarchical tree or graph structure for documents, but also add more complex facilities by associating each relation with a type of connective, employing a hierarchical label taxonomy, and allowing relations to connect discontinuous/overlapping segments.

Despite progress on new datasets in the frameworks listed above and many refinements to their guidelines, comparatively little progress has been made on discourse relation representation since the publication of PDTB. Because a full survey of the literature on computationally implementable theories of discourse relations and discourse organization is unfeasible in the scope of the current article,\footnote{See \citet{Stede2012} for an in depth overview.} we focus here on a synopsis and comparison of the main formalisms used in the field, for which substantial annotated corpus data exists: RST, SDRT, PDTB, CCR, and Discourse Dependencies.

\subsection{Rhetorical Structure Theory}\label{subsec:rst}

RST covers the most languages and datasets for discourse relations (12/26 datasets and 9/13 languages in the recent cross-formalism DISRPT shared task came from RST data, \citealt{braud-etal-2023-disrpt,braud-etal-2024-disrpt}). The theory distinguishes itself from other frameworks in its strong assumption of a tree constraint on all graphs, which must cover the entire text of a document, and the distinction of satellite vs. nucleus nodes (cf. Figure \labelcref{fig:gum-rst-example}). 

\begin{figure}[h!tb]

\includegraphics[clip,width=0.88\textwidth,trim = 0cm 0cm 0cm 0cm]{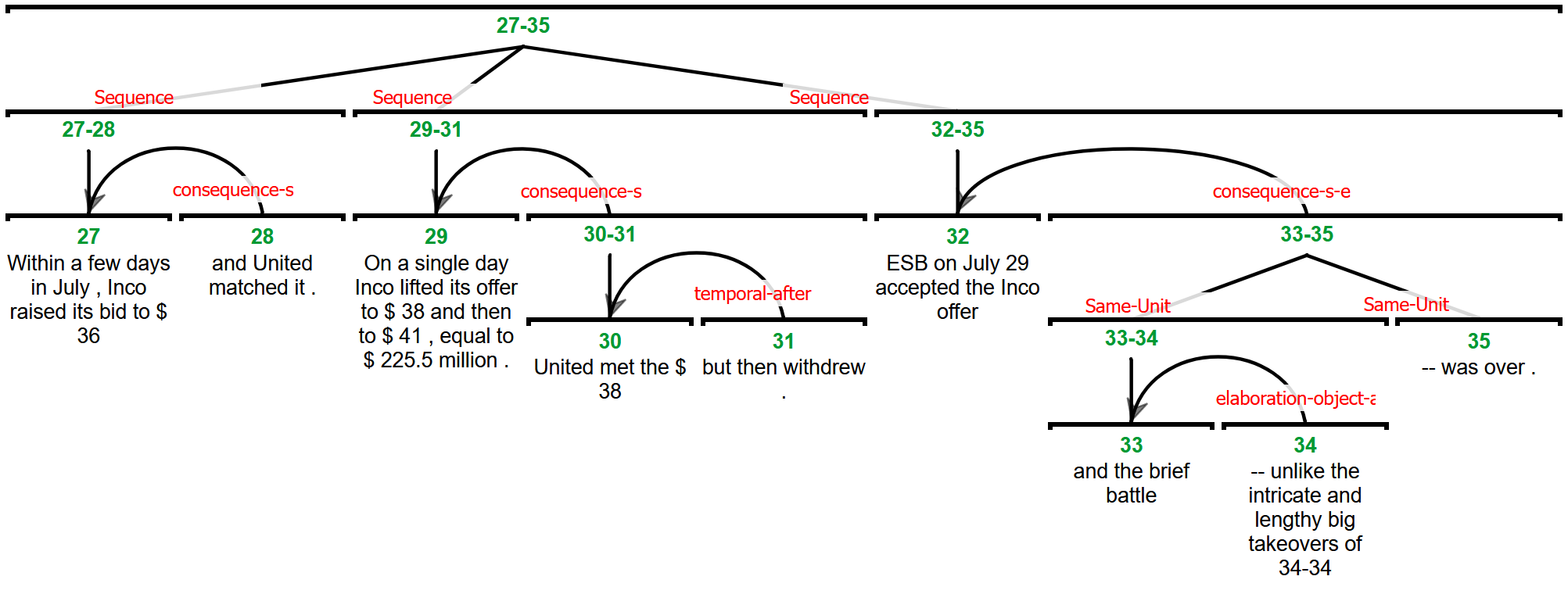}%
\centering
\caption{RST fragment from RST-DT: Satellites point to nuclei (e.g.~[28] is a \textsc{consequence} of [27]) while the symmetrical \textsc{sequence} relation connects equally prominent nodes. [33] and [35] form a discontinuous \textsc{same-unit}.}\label{fig:rstdt-ex}
\vspace{-7pt}
\end{figure}

The first large scale implementation of RST was the RST Discourse Treebank, annotating news wire material from the Wall Street Journal (WSJ) corpus \cite{MarcusSantoriniMarcinkiewicz1993}, with over 200K tokens in $385$ documents, and one of the largest inventories of relations ever implemented, with $78$ relations,\footnote{These include subtypes and variants accounting for different nuclearity patterns, which are often collapsed into $16$ coarse classes in automatic discourse parsing work, cf.~\citet[6]{HernaultPrendingerEtAl2010}.} as well as a pseudo-relation type called \textsc{same-unit}, used to connect parts of discontinuous units, as shown in [33-35] in Figure \labelcref{fig:rstdt-ex}.

\vspace{-2pt}

The figure also demonstrates two shortcomings of RST, which fed into the development of subsequent work. The first is that upon closer inspection, we may notice discourse relations that are unexpressed in the tree: Unit [31] contains two discourse markers explicitly signaling different relations. The word `then' corresponds to the annotated relation \textsc{temporal-after}, while the word `but' corresponds to no relation in the tree, but probably indicates the existence of a concurrent \textsc{concession} relation (see \citealt{moore-pollack-1992-problem}). The second shortcoming is the lack of a distinction between such explicitly marked relations, for which we can supply simple textual evidence as a rationale (e.g.~the existence of `then'), and implicit ones, such as the \textsc{consequence} satellite relation in [30-31], which is not indicated by a word like `then' or `but'.

A first attempt to address the latter shortcoming in RST was undertaken in the RST Signaling Corpus (\citealt{das2018rst}, RST-SC), which added signal type annotations to relations in the English RST-DT corpus, but did not anchor them to tokens. Thus for [31], the presence of explicit marking was annotated, but the word `then' was not identified as its locus. \citet{liu2019discourse} presented a pilot study on anchored signals for RST-DT, which was extended to four genres from an early version of the GUM RST treebank \cite{Zeldes2017}, anchoring signals to specific tokens \cite{liu-2019-beyond} -- the present work develops this idea further in Section \labelcref{sec:formalism} below.

\subsection{Segmented Discourse Representation Theory}\label{subsec:sdrt}

Segmented Discourse Representation Theory (SDRT, \citealt{AsherLascarides2003}), is the most similar framework to RST in assuming graphs covering entire documents, and discourse units connected recursively using relations defined independently of formal marking. As in RST, EDUs also coalesce to form complex discourse units, which are in turn joined with others to create larger units. SDRT is also notable in producing resources which explore discourse structure for multiparty dialogue, such as the STAC corpus \cite{asher-etal-2016-discourse} and the Molweni corpus \cite{li-etal-2020-molweni}, which focus on multiparty chat as part of an online game and in Ubuntu chat forums respectively.

\begin{figure}[b!ht]

\includegraphics[clip,width=0.9\textwidth,trim = 0cm 5cm 0cm 15cm]{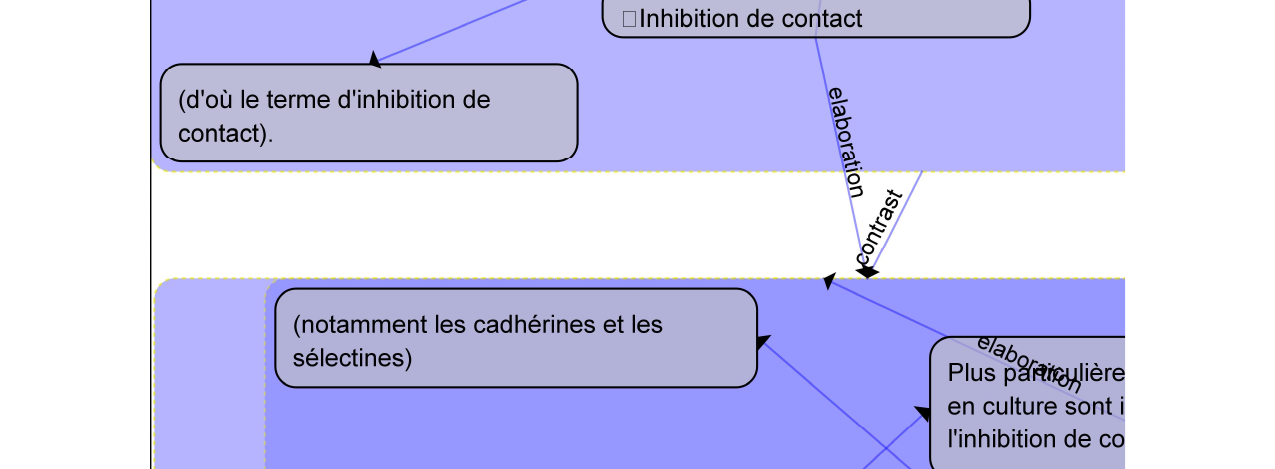}%
\centering
\caption{Fragment of an SDRT graph in the Glozz tool. The large blue discourse unit on the bottom has two incoming relations, \textsc{contrast} from the large blue unit at the top, and  \textsc{elaboration} from the gray EDU with the text `Inhibition de contact' at the top.}\label{fig:glozz}

\end{figure}

However, several differences distinguish SDRT, which also aligns with a specific formal semantic representation (DRT, \citealt{kamp2011discourse}) and defines relations as part of a formal logic. Notably, SDRT allows multiple relations between units, as in example \labelcref{ex:sdrt} from \citet{LascaridesAsher2007} and non-projective graphs, as shown in Figure \labelcref{fig:glozz} using Glozz \cite{WidloecherMathet2012}, the most commonly used interface for SDRT annotation.

\ex. ${\pi}_1$: John bought an apartment. ${\pi}_2$: But he rented it.\label{ex:sdrt} 

In \labelcref{ex:sdrt}, \citeauthor{LascaridesAsher2007} posit that unit ${\pi}_2$ forms both a \textsc{narration} relation and a \textsc{contrast} relation to ${\pi}_1$. SDRT also distinguishes coordinating relations, such as \textsc{contrast} from subordinating ones, such as \textsc{elaboration} \cite{AsherVieu2005}, but both can occur concurrently, as in Figure \labelcref{fig:glozz} for a French text from the ANNODIS corpus \cite{AfantenosDenisMullerEtAl2010}: The bottom complex unit (in blue) has incoming  \textsc{elaboration} and \textsc{contrast} relations, one from an EDU (in gray) and one from another complex unit. SDRT relations therefore do not reflect an RST-like notion of nuclearity or prominence. Units are also allowed to nest in each other, further complicating the data model.

\vspace{-5pt}

\subsection{Penn Discourse Treebank}\label{subsec:pdtb}

\begin{figure}[h!tb]

\includegraphics[clip,width=0.94\textwidth,trim = 0cm 0cm 0cm 0cm]{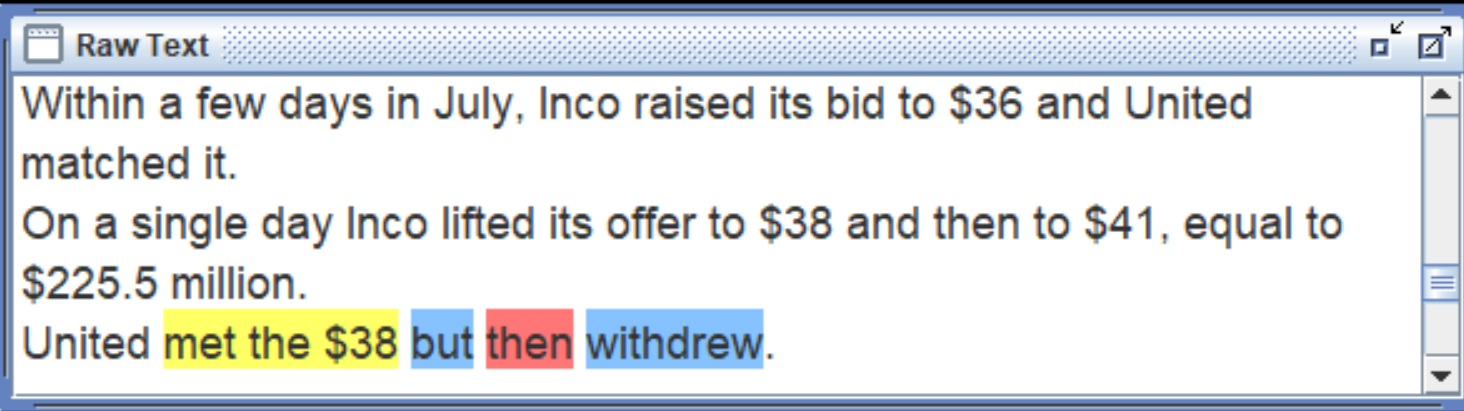}%
\centering
\caption{PDTB annotation interface for the same fragment from Figure \labelcref{fig:rstdt-ex}. Two concurrent relations are recognized, corresponding to \textit{but} and \textit{then} respectively.}\label{fig:pdtb-ex}
\vspace{-10pt}
\end{figure}

The Penn Discourse Treebank (PDTB) adopts a `lexically grounded' approach where discourse relations are annotated as senses of their associated discourse connectives \cite{prasad-etal-2014-reflections}. For instance, Figure \labelcref{fig:pdtb-ex} shows the same two concurrent relations from Figure \labelcref{fig:rstdt-ex}, identified by the two connectives, \textit{but} and \textit{then}: \textsc{Comparison.Concession.Arg2-as-denier} and \textsc{Temporal.Asynchronous.Precedence}. Such relations are called \textbf{explicit} relations in PDTB-style corpora. On the other hand, since there is no connective between the first two sentences in the figure, there is no explicit relation annotation. However, an implicit connective \textit{then} can be inserted between the two sentences (``\textit{...Inco raised its bid... \textbf{Then} on a single day Inco lifted...}''), and therefore an \textbf{implicit} relation instance is identified and annotated as \textsc{Temporal.Asynchronous.Precedence}. 

In addition to \textbf{explicit} and \textbf{implicit} relations, several other types are recognized in the English PDTB v3.0: 1) Alternative Lexicalizations (AltLex), AltLexC, EntRel, Hypophora, and NoRel \cite{prasad-etal-2014-reflections}. AltLex items are expressions not considered to be connectives by PDTB's syntactic guidelines, which limit connectives to subordinating or coordinating conjunctions, prepositional phrases, and adverbs (for example `let alone', which can mark an \textsc{Expansion.Conjunction}). 
AltLexC items are lexico-syntactic constructions which indicate relations, such as inverted auxiliaries marking a \textsc{Contingency.Condition} (e.g~``\textit{Had I done it...}'').

Arguments associated with identified relations in PDTB follow the principle of minimality: Only the minimal text needed for a given relation will be selected which can be sentences, clauses, nominalizations, verb phrases etc. \cite{prasad-etal-2014-reflections}. Additional text that is relevant but not necessary for the interpretation can be selected as supplementary information during annotation. 

A major shortcoming of PDTB is the lack of higher-level structure over the relations between text spans (compare this with the RST annotation of the same fragment in Figure \labelcref{fig:rstdt-ex}, which constructs an overarching nested structure). However, the lack of high-level structure makes annotation easier compared to RST and SDRT, as high-level structures are considered more challenging \cite{Peng2023}. Thus, PDTB has allowed the creation of large corpora in a variety of languages such as Chinese \cite{CDTB-LDC}, Turkish \cite{zeyrek-kurfali-2017-tdb}, Portuguese \cite{CRPC-DB-Portuguese}, and Italian \cite{tonelli-etal-2010-annotation} as well as for multilingual versions of TED talks (TED-MDB, \citealt{zeyrek2018multilingual,zeyrek2019ted}).

\subsection{Cognitive Approach to Coherence Relations}\label{subsec:ccr}

Cognitive Approach to Coherence Relations (CCR, \citealt{SandersSpoorenNoordman1992}), unlike most other discourse frameworks, offers a psycholinguistic account of discourse relations and discourse signalling, focusing on discourse comprehension. CCR defines discourse comprehension as a psychological mechanism that creates a coherent representation of the text content based on the ways text segments are linked with each other by discourse relations. CCR characterizes relations as a configuration of five dimensions, each decomposed into binary attributes \cite{Sanders-et-al:2021}:

\begin{itemize}
    \item Polarity: positive or negative discourse relations
    \item Basic operation: causal (strongly linked) or additive (weakly linked) relations
    \item Source of coherence: objective (semantic) or subjective (pragmatic) relations\footnote{Similarly, RST relations are sometimes classified as either \emph{subject matter} or \emph{presentational} relations.}
    \item Implication order: basic (antecedent-consequent) or non-basic (reverse) order
    \item Temporality: temporal or non-temporal relations
\end{itemize}

As an example, consider the relation in \labelcref{CCR:example}, from \citet[11]{Sanders-et-al:2021}, which CCR decomposes as follows: The relation expresses a denial of expectation,\footnote{For a \emph{positive causal} relation, the implication would be: police not involved $\rightarrow$ no need to fear reprisals.} and hence, represents a \emph{negative causal} relation ({\sc concession} in RST). The relation links two segments that express facts; so, it is an \emph{objective} relation. The implication order is \emph{basic} since the antecedent segment precedes the consequent segment. Furthermore, the linear sequence of the segments represents a chronological progression, which makes the relation a \emph{temporal} one.

\ex. Although [they were officially assured the police would not be involved in the census] [many people are afraid of reprisals ... ]\label{CCR:example}

CCR considers discourse signals (connectives/cue phrases) as processing instructions guiding the reader to infer the relation between segments \cite{Sanders-et-al:2007}.  
In the absence of such signals, CCR postulates that relation identification may require additional cognitive energy and longer processing time, affecting text comprehension. Evidence to support these claims comes from both psycholinguistic and corpus-based studies (see \citealt{Kleijn-et-al:2019,Sanders-et-al:2021}). CCR annotation, like PDTB's, targets local-level relations and their connectives. CCR corpora, albeit fewer in number, are available for English \cite{rehbein-et-al:2016} and Dutch \cite{Vis-et-al:2012}.

\subsection{Discourse Dependency Structure}\label{subsec:deps}

Discourse Dependency Structure (DDS, \citealt{li-etal-2014-text, morey-etal-2018-dependency}) deviates from RST's constituency structure (a.k.a.~\textit{c-tree}) and connects EDUs using binary and asymmetrical dependency relations to facilitate parsing (i.e.~\textit{d-trees}). 
DDS aligns with widely-used syntactic dependency structures such as Universal Dependencies (UD, \citealt{nivre-etal-2020-universal}) and offers a simple and transparent tree structure for annotating document-level discourse relations \cite{morey-etal-2018-dependency}. 

\begin{figure}
    \centering
    \includegraphics[clip,width=0.95\textwidth,trim = 0cm 22.3cm 0cm 4.3cm]{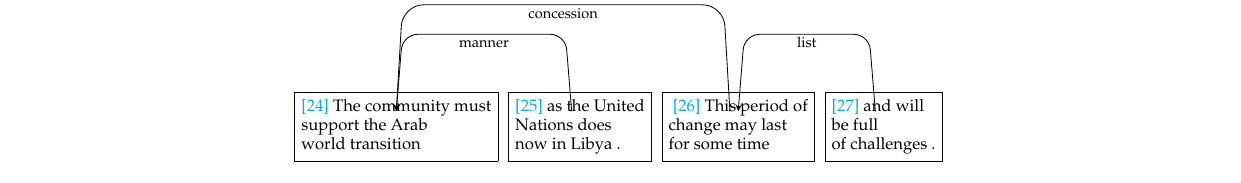}

\caption{Head-ordered DDS converted from the RST fragment in \Cref{fig:gum-rst-example}.}
\label{fig:gum-rst-dependencies-example}
\vspace{-22pt}
\end{figure}

Only a few discourse treebanks are annotated natively in DDS, including SciDTB \citep{yang-li-2018-scidtb}, SciCDTB \cite{cheng-li-2019-zero}, and COVID19-DTB \citep{nishida-matsumoto-2022-domain}.
Most DDS data is converted from corpora in other frameworks, e.g.~\citet{hirao-etal-2013-single} and \citet{li-etal-2014-text} designed transformations from RST-DT to obtain parent-child relations between EDUs for summarization and discourse parsing. Both approaches produce binarized, asymmetric dependency trees translating nuclearity to headedness, while differing in the handling of multinuclear relations.
\citet{morey-etal-2018-dependency} further add head-ordering to preserve the scope of satellite modifications and render conversions between constituent and dependency trees bijective. 
\Cref{fig:gum-rst-dependencies-example} presents a converted head-ordered DDS equivalent to the fragment in \Cref{fig:gum-rst-example}. The multinuclear \textit{list} relation is transformed into a right-to-left dependency arc, with a chain modifying the first EDU in [24]. 

DDS datasets have also been converted from SDRT and PDTB data, with the latter complemented by automatically predicted higher-level relations \cite{stede-etal-2016-parallel, yi-etal-2021-unifying}. 
Due to the lack of large-scale DDS-native treebanks, discourse dependency parsing models are either trained on converted datasets \citep{yi-etal-2021-unifying} or through zero/few-shot learning and bootstrapping \citep{cheng-li-2019-zero, nishida-matsumoto-2022-domain}.

\subsection{Multiple Frameworks}\label{subsec:multi}

Some multilayer datasets have been developed which contain analyses in multiple frameworks in parallel. RST-DT and PDTB contain overlapping material from the WSJ corpus, allowing for some framework comparisons \cite{DembergEtAl2019}. \citet{StedeNeumann2014} and \citet{bourgonje-stede-2020-potsdam} added connective annotations for explicit relations to the German RST-annotated Potsdam Commentary Corpus (PCC). \citet{SunWang2022} constructed a corpus of $500$ Chinese `run-on' sentences annotated with both RST and PDTB-style analyses. However to the best of our knowledge, this article is the first attempt at producing a new theory incorporating insights and advantages from the multiple frameworks described above, in which complete RST-style trees with nuclearity are anchored to connectives and other signals, while allowing tree-breaking relations as postulated in SDRT. The next section defines the scope of the formalism, before presenting data and parsing experiments implementing our analyses.

\section{Formalism}\label{sec:formalism}

Analyses in \theory{} aim to retain the benefits of RST trees (nuclearity and its applications to recursive summarization and information extraction, dialogue planning, etc.), while addressing weaknesses discussed in Section \labelcref{subsec:rst}. Analyses consist of $3$ components which we discuss below:

\begin{enumerate}
    \item A primary, single-rooted, labeled projective n-ary constituent tree over non-overlapping Elementary Discourse Units, which cover the text
    \item A possibly empty set of secondary labeled, directed and possibly cyclic and/or non-projective edges, which are licensed under specific conditions
    \item A possibly empty set of categorized signals associated with a set of tokens and a single primary or secondary edge from [1] or [2]
\end{enumerate}

Although in the study below we expand on a corpus with existing primary trees, the formalism is intended to be applicable to the analysis of plain text, for which a primary tree would then be prepared as part of the \theory{} analysis.

\subsection{The Primary Tree}\label{sec:primary-tree}

A primary tree $G$ 
is defined, as in traditional RST, as a directed, single-rooted and fully connected labeled tree. 
Let $V$ be a set of terminal and non-terminal vertices with a subset of ordered terminals $S$ which are segments covering the tokens of the text $T$, and edges $E$ between vertices with labels from the set $L$:
\vspace{-10pt}

\begin{equation}
\begin{aligned}
    G = & \text{ $\langle V, S, E, L, T\rangle $} \\
    V = & \text{ $\{v_1, v_2, ..., v_n\}$} \\
    S \subseteq &\ V \\
    E = & \text{ $\{\langle v_i, v_j \rangle \mid v_i \in V \land v_j \in V \land v_i$ is the parent of $v_j\}$ } \\
    L = & \text{ $\{l_1, l_2, ..., l_m\}$} \\
    T = & \text{ $\{t_1, t_2, ..., t_k\}$} \\
\end{aligned}
\end{equation}

Note that $S$ is actually in almost all cases a \textit{proper} subset of $V$: The sole exception is the degenerate case where there is only a single EDU, which produces $S = V$. 
The tree is further constrained to be projective. 
All tokens belong to exactly one terminal segment (i.e.~EDU), and there is a single unique label for each edge:
\begin{equation}
    \begin{aligned}
    \forall\ t  \in T [ \exists!\ s \in S &[ s \text{ contains } t ]] \\
    \forall\ v  \in V [ \exists!\ l \in L &[ l  \text{ labels }  v ]] 
    \end{aligned}
\end{equation}

In addition, each node in $V$ is classified as a satellite or nucleus node, and for each non-terminal node in $V$, at least one child node is a nucleus. 
\theory{} allows $n$-ary branching trees, though binarization via Chomsky Normal Form (CNF) is possible as a trivial conversion for use with binary parsers.

The criteria for building trees are the same as in RST (see \citealt{MannThompson1988,TaboadaMann2006}), and will not be discussed in depth due to space. Briefly, propositions are grouped together based on the function they serve, with more prominent or necessary units being assigned the nucleus status, and less necessary or omittable units serving as satellites. Labels are defined based on relations' effect on the reader or hearer, and are assigned based on the perceived intention of the author or speaker to have such an effect, e.g.~a group of EDUs which is perceived as an explanation supplying evidence for a claim in another group of EDUs may be analyzed as a satellite to the latter group, with a label such as \textsc{explanation-evidence}. While \theory{} as a theory does not necessarily prescribe a specific set of relation labels, the labels in this article will come from the inventory of the GUM corpus, which has $32$ total labels, including the label \textsc{same-unit} to connect parts of discontinuous EDUs (see Appendix \hyperref[app:rels]{A} for the full inventory).

\begin{figure}[h!tb]
    \centering
    \begin{subfigure}[t]{0.45\textwidth}
        \centering
        \includegraphics[width=\linewidth]{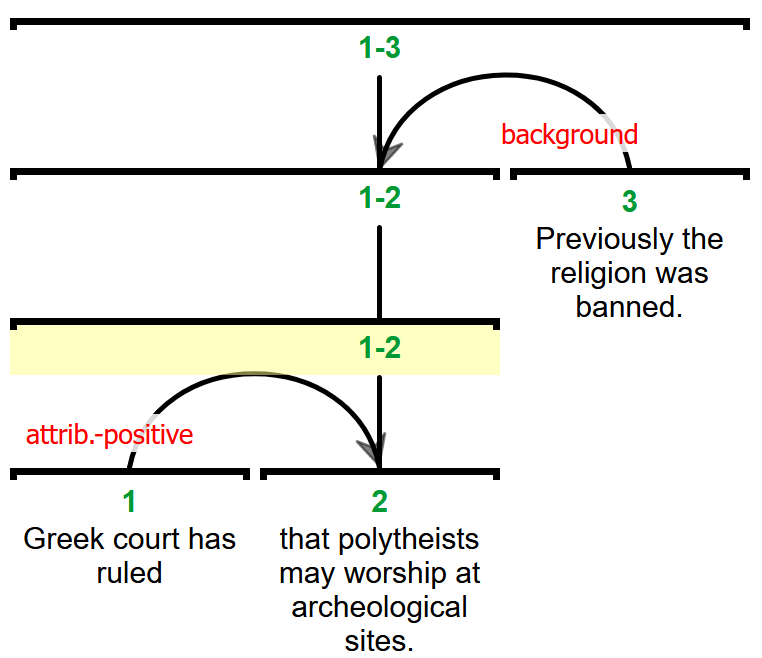} 
        \caption{`no empty hierarchy'} 
    \end{subfigure}
    \hfill
    \begin{subfigure}[t]{0.45\textwidth}
        \centering
        \includegraphics[width=\linewidth]{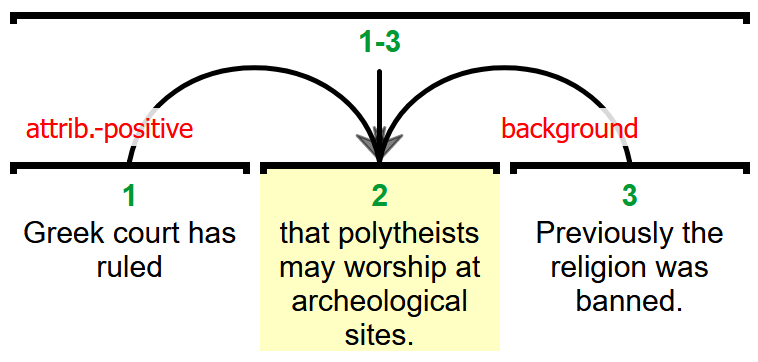} 
        \caption{`no satellite ties'}
    \end{subfigure}

\caption{Violations of Additional Constraints in \theory{}.}
\label{fig:violations}
\vspace{-10pt}
\end{figure}

\theory{} primary trees are thus largely identical to RST trees, with some constraints made more explicit than in previous implementations. Specifically, \theory{} trees must define an explicit word \textit{tokenization} of EDU contents to allow for the alignment of signals (see Section \labelcref{sec:signals} below); it is assumed that there is no empty hierarchy, i.e.~each non-terminal has at least two children, and hierarchy is strictly ordered without ties. These constraints mean that there are no unary derivations, and no two satellite children for the same node. Violations of both constraints are illustrated in Figure \labelcref{fig:violations}.\footnote{These constraints are often applied in RST trees in practice, but are occasionally violated in most datasets, and would cause problems for some of the algorithms we use for aligning signals below.} 

On the left, a redundant span has no child relations (the lowest span labeled [1--2]); on the right, [2] has two satellites which are not hierarchically ordered -- instead \theory{} requires that [1] scopes over [2--3] (meaning the \textsc{attribution} contents of what the court ruled include the other two units, or that [3] scopes over [1--2], forming \textsc{background} to both. In this case, the latter option is the correct one.

\subsection{Secondary Edges}

As noted earlier, some discourse relations occur in texts which cannot be expressed in a projective, acyclic tree as defined in \labelcref{sec:primary-tree}. To represent such relations, we define a subset of edges, called \textbf{secondary edges}, which are not constrained by limitations on projectivity or cycles.\footnote{The term `secondary edge' is inspired by the `secedges' in the German Tiger Treebank \cite{BrantsDipperHansenEtAl2002}, where similar edges where added to a primary syntax tree for tree-breaking dependencies.} Secondary edges are permitted to connect any two nodes in the primary tree, including nodes which are already connected by a primary edge, subject to the following constraints. A secondary edge:
\begin{enumerate}
    \item may only be added if it is supported by a sufficient \textbf{signal}
    \item may only connect two nodes which are not already connected by a secondary edge with the same directed path
    \item may not connect a node to itself
    \item may not require the definition of additional nodes
\end{enumerate}
Constraints [2--3] mean that any two nodes $v_1, v_2$ in the tree can, at most, be connected via three edges: A primary edge, a secondary edge $v_1 \rightarrow v_2$ and a secondary edge $v_2 \rightarrow v_1$. This places an upper bound on the complexity of the formalism (see Section \labelcref{sec:complexity}).

\begin{figure}[h!tb]
    \centering
    \begin{subfigure}[t]{0.49\textwidth}
        \centering
        \includegraphics[width=\linewidth]{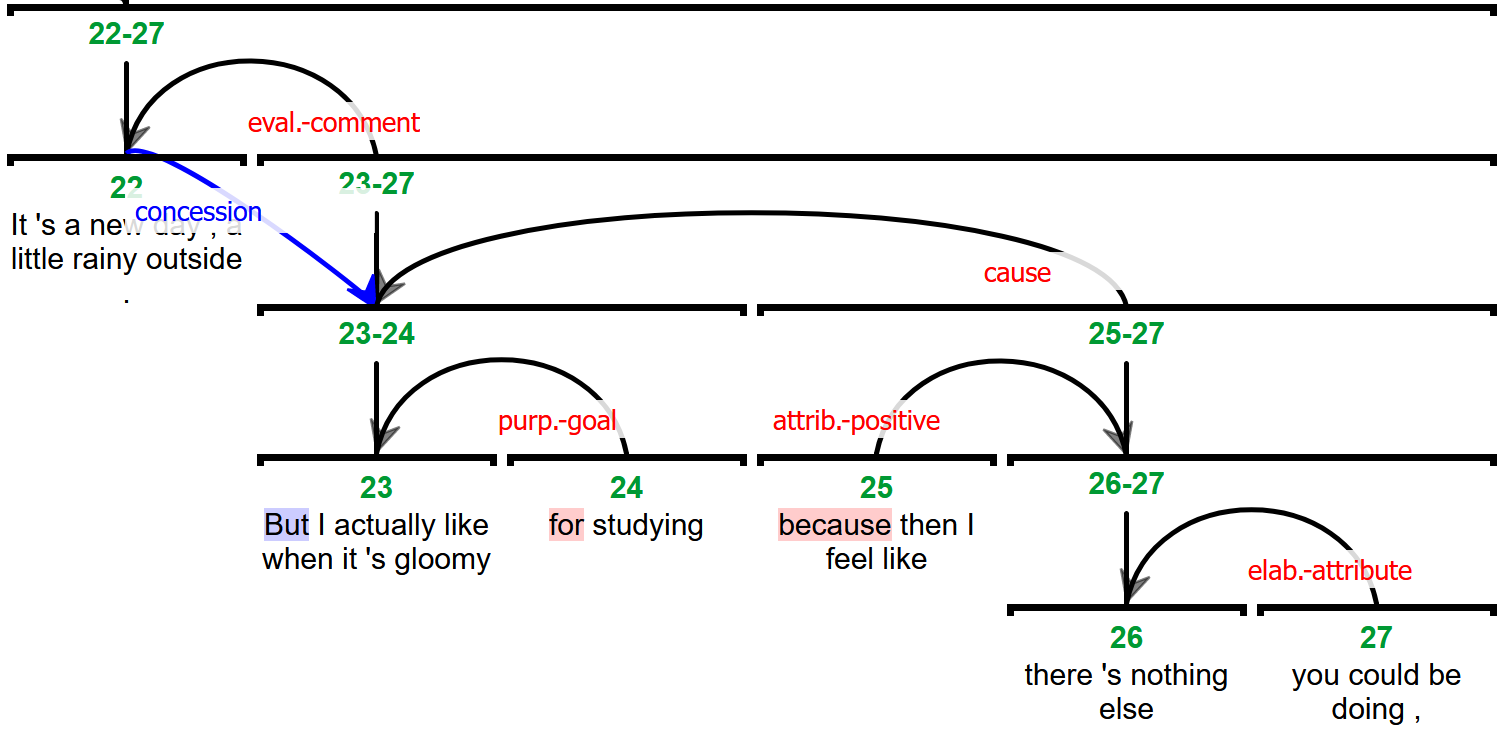} 
        \caption{Secondary \textsc{concession} with orphan `but'} 
    \end{subfigure}
    \hfill
    \begin{subfigure}[t]{0.49\textwidth}
        \centering
        \includegraphics[width=\linewidth]{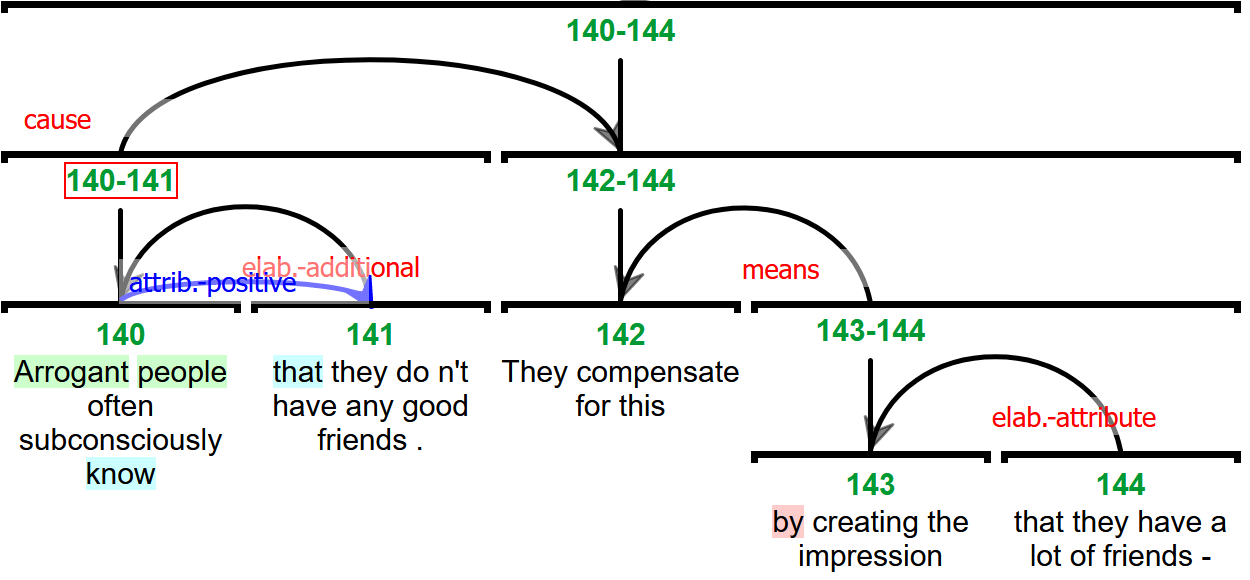} 
        \caption{Secondary \textsc{attribution}}
    \end{subfigure}

\caption{Secondary edges licensed by an orphan DM and reported speech. On the left, a secondary \textsc{concession} in blue points from 22 to the sentence containing 23--24; on the right, a secondary positive \textsc{attribution} points from 140 to 141.}
\label{fig:secedges}
\vspace{-6pt}
\end{figure}

Constraint [1] lies at the core of our proposal for additional relations: Since agreement on discourse relations is already challenging, we want to limit additional edges only to clearly signaled cases. While it is conceivable that a variety of definitions could be used for `clear signals', we limit our proposal to two kinds of signals: Discourse markers (DMs) like `but' or `then'\footnote{We use the term `discourse marker' with largely the same definition and items as PDTB connectives, but with the difference that the spans they connect correspond to RST EDUs or complex discourse units, rather than argument spans following PDTB guidelines.} for which no corresponding associated parent relation can be found, which we refer to as \textbf{`orphan DMs'}; and unambiguous morphosyntactic signals, in our implementation specifically restricted to either \textbf{reported speech} which is not already captured in a primary \textsc{attribution} relation,\footnote{We note that \textsc{attribution} is especially well known to co-exist alongside other relations \cite{potter-2019-rhetorical}, and has merited concurrent treatment in PDTB as well.} and adnominal clauses which are not already captured using an embedded adnominal \textsc{elaboration} or \textsc{purpose} relation, e.g.~relative clauses not interpreted as a primary \textsc{elaboration}.\footnote{We do not rule out that other types of reliable signals could be added to license further secondary edges in future work.} Figure \labelcref{fig:secedges} demonstrates these two types of licensing conditions.

On the left, the secondary edge captures the relation corresponding to the orphan DM `but' (highlighted in dark blue), which has no corresponding relation in the primary tree -- instead, the annotator perceives the main function of [23--27] as an \textsc{evaluation} of how a rainy day isn't too bad; note also that the nuclearity of the \textsc{evaluation} goes in the opposite direction of the \textsc{concession}, and that they do not scope over the same part of the text (the secondary edge connects [27] with [23--24]). 

On the right, the fact that arrogant people subconsciously know they have no friends ([140--141]) is seen as the \textsc{cause} of compensating for this. Although `know' is a typical \textsc{attribution} verb, the annotator has analyzed the nucleus of the causal predicate to be \textit{knowing} that they have no friends ([140]), rather than the fact that they have no friends in itself ([141]), forcing them to make [141] an \textsc{elaboration} to [140]. Nevertheless, syntactically [140--141] unambiguously follows the reported speech/cognition verb pattern, licensing a secondary \textsc{attribution} signaled by the predicate `know' and the complementizer `that' (syntactic signals highlighted in cyan), and the attribution source `Arrogant people', a semantic signal highlighted in green.

As shown in these examples, the relation inventory for secondary edges is assumed to be the same set of labels L used for the primary tree, though we note that if the inventory distinguishes a multinuclear and a satellite version of the same relation, these collapse and become indistinguishable for secondary edges, which do not indicate nuclearity by nature (though they do indicate directionality -- for example a secondary \textsc{concession} still has a conceded part and a claim the concession contrasts with). In this way, \theory{} offers a partial remedy to previous criticism of RST in situations in which a single nuclearity choice may not express everything we want to know about discourse structure \cite{moore-pollack-1992-problem,Stede2008a}, by allowing the expression of concurrent relations; however, the cost of retaining the advantages of the primary tree is that nuclearity itself is still kept unambiguous -- secondary edges express only relations, and not overall prominence in the discourse structure.

As an annotation practice, we therefore recommend that secondary edges should only be added after the complete primary tree has been annotated, so that the most prominent relations can determine nuclearity without considering the presence of orphaned DMs. The first example in Figure \labelcref{fig:secedges} illustrates why this is important: If primary and secondary edges are annotated concurrently, annotators may be tempted to select the unmarked relation as primary, and utilize the resulting orphan `but' to establish the second relation, so that both can be marked. If the relation not corresponding to a DM is deemed more prominent from a functional perspective, as in Figure \labelcref{fig:secedges}(a), then this is what we want; but that means the primary relation must always be established first, or else we may compromise our standards for determining nuclearity. In other words, the primary tree in \theory{} should be the same tree as in plain RST.

Finally we note that not all potential DM items will necessarily receive a corresponding edge. Although we may expect any DM without a corresponding primary edge to automatically receive a secondary one, this will be ruled out in two cases: 1. when there are two orphan DMs which can be associated with secondary edges along the same path, due to constraint [2]; and 2. when the necessary spans for the edge to connect do not exist, conflicting with constraint [4]. While we did not encounter the first situation in our data (see Section \labelcref{sec:annotation}), the second issue has occurred, especially when the necessary spans do not exist due to segmentation guidelines, as in \labelcref{ex:trapped-orphan}. 

\ex. $[$\textbf{If} you live in or near a big city,$]_{<condition>}$ it is easier to attract enough customers .. $[$\textbf{than} \textbf{if} you live in a sparsely populated rural area .$]_{<antithesis>}$\label{ex:trapped-orphan}

In this example, there are three subordinating conjunctions: Two `if's and one `than'. The first `if' marks a primary \textsc{condition} relation, and `than' marks an \textsc{antithesis}. Although the second `if' clearly has a conditional sense, it is not a condition of the main clause (`easier to attract customers'), but rather a condition for the implied elliptical clause that might have followed `than' (`easier .. than \textit{it would be}'). However, because such a clause does not appear, EDU segmentation guidelines prevent the existence of the necessary argument span for a secondary relation, and the orphaned second `if' remains without a corresponding edge.

\subsection{Signals}\label{sec:signals}

Like PDTB, \theory{} anchors relations to markers in a text called `signals' \cite{liu-2019-beyond}: This allows one not only to know which relation is indicated by which signal(s), but also to pinpoint exactly which words/phrases/constructions in the text contribute to the signalling of the relation. However, we assume a broader perspective on signalling than PDTB, encompassing much more than DMs and similar expressions. Following the taxonomy proposed by RST-SC \cite{RST-SC-annotation-manual}, we divide non-DM signals into seven types, corresponding to: \emph{graphical}, \emph{lexical}, \emph{morphological}, \emph{numerical}, \emph{reference}, \emph{semantic}, and \emph{syntactic} features.

These types are divided into further subtypes, shown with examples in Table \labelcref{tab:sig-types}. For example, the \emph{reference} type indicates that cohesion is signaled by anaphoric reference to an entity, with four subtypes: \emph{personal} (anchored to a personal pronoun and its antecedent), \emph{demonstrative} (an anaphoric NP headed or determined by a demonstrative), \emph{comparative} (a comparative or relative expression, e.g.~`(an)other' anaphora), and \emph{propositional} reference (e.g.~NPs referring back to a verbal phrase). Some signal types are anchored to a lexical indicative expression such as a word (e.g.~`nice' signaling an \textsc{evaluation}), phrase, or other alternate expression, with the latter corresponding to PDTB's inventory of AltLex signals. Other signals may be anchored to tokens which are only signals in very specific contexts, such as negations accompanying an adversative relation with the same predicate in positive and negative environments, which can be crucial to interpreting contrasts (cf.~\citealt{webber:2013}). 
Others still refer to paired tokens, such as brackets or quotation marks, while some are not anchored to any tokens, such as graphical layout (a heading identified by its font, size, and separate appearance), or placement in a sequence of indented blocks or bullet points.\footnote{Following acceptance of this paper, we have also been discussing the possibility of incorporating implicit connectives, following PDTB definitions, as a type of signal not anchored to specific tokens, or even as a possible trigger for another type of secondary edges. We leave this idea for exploration in further work.}

\begin{table}[h!tb]
\centering
    \resizebox{\textwidth}{!}{%
\midsepremove
\begin{tabular}{l|l|l}
\toprule
\textbf{signal type} & \textbf{subtypes} &  \textbf{example}  \\
\midrule
\rowcolor{LightYellow}
graphical & colon, dash, semicolon & \textit{$[$Let me tell you a story \textcolor{red}{\textbf{:}}$]$}$_{<organization-preparation>}$  \\
\rowcolor{LightYellow}
 & layout & \textit{$[$Introduction$]$}$_{<organization-heading>}$  \\
 \rowcolor{LightYellow}
 & items in sequence & \textit{1. wash $[$2. cut$]$}$_{<joint-list>}$  \\
\rowcolor{LightYellow}
 & parentheses, quotation marks & \textit{it rained $[$\textcolor{red}{\textbf{(}}and snowed a bit\textcolor{red}{\textbf{)}}$]$}$_{<elaboration-additional>}$  \\
\rowcolor{LightYellow}
 & question mark & \textit{$[$Did you\textcolor{red}{\textbf{?}}$]$}$_{<topic-question>}$\textit{ No.}  \\
\midrule
\rowcolor{LightOrange}
lexical & alternate expression & \textit{He agreed. $[$\textcolor{red}{\textbf{That is}} he said yes$]$}$_{<restatement-repetition>}$  \\
\rowcolor{LightOrange}
 & indicative word/phrase & \textit{They planned a party! $[$That's \textcolor{red}{\textbf{nice}}/\textcolor{red}{\textbf{Can't wait}}!$]$}$_{<evaluation-comment>}$  \\
 \midrule
\rowcolor{LightTan}
morphological & mood & \textit{\textcolor{red}{\textbf{Go}} with them $[$I think you should$]$}$_{<explanation-motivation>}$  \\
\rowcolor{LightTan}
& tense & \textit{I \textcolor{red}{\textbf{started}} an hour ago, $[$now I\textcolor{red}{\textbf{'m}} resting$]$}$_{<joint-sequence>}$  \\
\midrule
\rowcolor{LightPink}
numerical & same count & \textit{$[$\textcolor{red}{\textbf{Two}} reasons.$]$}$_{<organization-preparation>}$\textit{ First…}  \\
\midrule
\rowcolor{LightGrey}
reference & 
comparative & \textit{$[$I don’t want \textcolor{red}{\textbf{it}}$]$}$_{<adversative-antithesis>}$\textit{ I want \textcolor{red}{\textbf{another one}}.}  \\
\rowcolor{LightGrey}
& demonstrative / personal & \textit{They met \textcolor{red}{\textbf{Kim}}. $[$\textcolor{red}{\textbf{This person}} / \textcolor{red}{\textbf{she}} was…$]$}$_{<elaboration-additional>}$  \\
\rowcolor{LightGrey}
& propositional & \textit{\textcolor{red}{\textbf{They met Kim}}. $[$\textcolor{red}{\textbf{This encouner}} was…$]$}$_{<elaboration-additional>}$  \\
\midrule
\rowcolor{LightGreen}
semantic & antonymy & \textit{Beer is \textcolor{red}{\textbf{cheap}}, $[$wine is \textcolor{red}{\textbf{expensive}}$]$}$_{<adversative-contrast>}$  \\
\rowcolor{LightGreen}
& attribution source & \textit{$[$\textcolor{red}{\textbf{Kim}} said$]$}$_{<attribution-positive>}$\textit{ they would}  \\
\rowcolor{LightGreen}
& lexical chain & \textit{it was \textcolor{red}{\textbf{funny}} $[$so they \textcolor{red}{\textbf{laughed}}$]$}$_{<causal-result>}$  \\
\rowcolor{LightGreen}
& meronymy & \textit{\textcolor{red}{\textbf{The house}} was big, $[$\textcolor{red}{\textbf{the door}} two meters tall$]$}$_{<elaboration-additional>}$  \\
\rowcolor{LightGreen}
& negation & \textit{Kim danced, $[$Yun did\textcolor{red}{\textbf{n't}} dance$]$}$_{<adversative-contrast>}$  \\
\rowcolor{LightGreen}
& repetition/synonymy & \textit{They met \textcolor{red}{\textbf{Dr. Kim}}. $[$\textcolor{red}{\textbf{Dr. Kim}}/\textcolor{red}{\textbf{The surgeon}} was…$]$}$_{<elaboration-additional>}$  \\
\rowcolor{LightGreen}
\midrule
\rowcolor{LightCyan}
syntactic & infinitival/relative clause & \textit{a plan $[$\textcolor{red}{\textbf{to}} win$]$}$_{<purpose-attribute>}$  \\
\rowcolor{LightCyan}
& interrupted matrix clause & \textit{$[$I meant --$]$}$_{<orgnization-phatic>}$\textit{ I mean, }  \\
\rowcolor{LightCyan}
& modified head & \textit{a \textcolor{red}{\textbf{plan}} $[$to win$]$}$_{<purpose-attribute>}$  \\
\rowcolor{LightCyan}
& nominal modifier & \textit{articles $[$\textcolor{red}{\textbf{explaining}} chess$]$}$_{<elaboration-attribute>}$  \\
\rowcolor{LightCyan}
& parallel syntactic construction & \textit{\textcolor{red}{\textbf{it's all}} tasty $[$\textcolor{red}{\textbf{it's all}} pretty$]$}$_{<joint-list>}$  \\
\rowcolor{LightCyan}
& past/present participial clause & \textit{Kim appeared $[$\textcolor{red}{\textbf{dressed}} in black$]$}$_{<elaboration-attribute>}$  \\
\rowcolor{LightCyan}
\rowcolor{LightCyan}
& reported speech & \textit{$[$Kim said$]$}$_{<attribution-positive>}$\textit{ \textcolor{red}{\textbf{that}} they \textcolor{red}{\textbf{would}}}  \\
\rowcolor{LightCyan}
& subject auxiliary inversion & \textit{I would have $[$\textcolor{red}{\textbf{had}} I known$]$}$_{<contingency-condition>}$  \\
\bottomrule
\end{tabular}
\midsepdefault
}
\caption{Non-DM signal types and subtypes, with examples highlighting in red the signal tokens which indicate the relation of the unit in square brackets.}\label{tab:sig-types}
\vspace{-5pt}
\end{table}

\begin{figure}[H]
\centering
\includegraphics[width=0.7\textwidth]{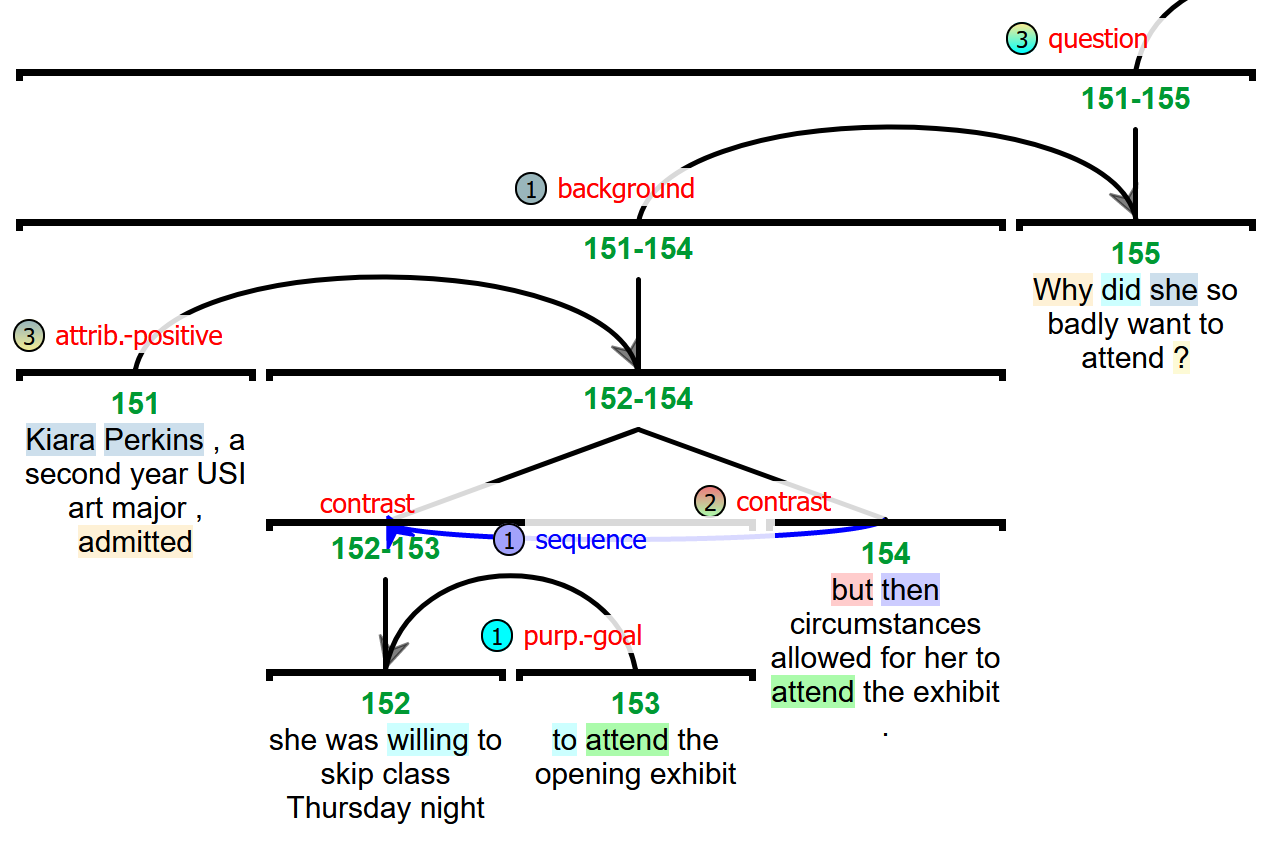}
\caption{A larger fragment annotated in \theory{}.}
\label{fig:rstpp-fragment}
\vspace{-8pt}
\end{figure}

Figure \labelcref{fig:rstpp-fragment} shows a larger \theory{} graph fragment, which differs from a corresponding basic RST tree only in the addition of highlighted signals (background colors for tokens) and tree-breaking secondary edges (blue arrow edge type). Units [151-154] function as \textsc{background} to a question `Why did she so badly want to attend?' ([155]), which carries $3$ signals (notice the number `3' next to the relation \textsc{question}): A \textit{lexical} signal in dark yellow (`Why'), a \textit{graphical} one in light yellow (the `?') and a syntactic auxiliary inversion (in cyan, anchored to `did' in [155]). The \textsc{background} relation has a personal \textit{reference} signal in gray (`Kiara Perkins ... she', indicating the background relates to this person), and the contents of the \textsc{background} is attributed to Perkins in [151], signaled by a \textit{semantic} \textit{attribution source} (another signal anchored to the span `Kiara Perkins'), an attribution verb `admitted' (\textit{lexical}, in yellow), and a complement clause headed by `willing' (\textit{syntactic, reported speech}, in cyan). The \textsc{purpose-goal} clause in [153] is signaled by a to-infinitive (\textit{syntactic, infinitival clause}, anchored to `to'), and the \textsc{contrast} is marked by `but' (a DM, in red) and the lexical chain `attend ...  attend'. Finally, an orphan `then' (in blue) indicates the presence of the secondary \textsc{sequence} relation. More details on the annotation interface used for the visualization  are given in Section \labelcref{sec:annotation}.

\subsection{Complexity and Effort}\label{sec:complexity}

Although \theory{} graphs as shown in Figure \labelcref{fig:rstpp-fragment} are quite complex, the computational complexity of derivations in \theory{} is not very much higher than in RST. Because we retain the premise of a primary tree, and we constrain secondary edges to connecting at most each pair of nodes in each direction once, parsing secondary edges in the worst case scenario adds a single step in quadratic time. As for signal detection, although each span of tokens can be used multiple times with multiple signal types or for multiple relations (see the example `Kiara Perkins' above, which serves as both an \textit{attribution source} and a \textit{personal reference} signal for different relations), the problem of signal detection can be approached as token-wise multilabel classification, where each token must be classified for a complex signal type (we can consider e.g.~reference:personal\_reference to be a single label) and a pointer to a corresponding relation from the closed set of relations available in the tree. In practice, automatic \theory{} parsing can be substantially less costly using a pipeline, as we will show in Section \labelcref{sec:parsing}, or potentially a joint model.

In terms of annotation effort, an anonymous reviewer has brought up the relatively high cost of a manual RST analysis and its extensions using \theory{}, while another reviewer has asked why the starting point of the analysis is RST-like, rather than PDTB-like. To adress the first point, we certainly agree that building primary trees is labor intensive, an aspect in which \theory{} does not differ from RST; however the amount of effort associated with the addition of secondary edges and signals can be reduced by relying on preprocessing from NLP tools for tasks which show promising performance, such as connective detection \cite{GesslerEtAl2021,liu-etal-2023-hits}, provided that these can be aligned to trees (see Section \labelcref{sec:results}). Unalignable signals could then be inspected as indicators that secondary edges may be needed. Although the cost for \theory{} annotation at the moment is high, we would argue that it is unavoidable for a computationally implemented theory accounting for discourse relations and the devices used to mark them, since these will inevitably include multiple, concurrent instances. We are hopeful that with increasing performance of NLP models, much of the task could be done automatically, and we give some first numbers for complete automatic \theory{} annotation in Section \labelcref{sec:results}.

Regarding the use of RST, rather than PDTB as a starting point, \theory{} focuses on incorporating two of PDTB’s greatest advantages into a hierarchical discourse representation: 1., the inclusion of multiple concurrent relations, and 2., providing a text-anchored rationale for relations (DMs and other signal types). However strictly building on top of PDTB as a framework would bring in its shortcomings (see Section \labelcref{subsec:pdtb}): lack of the usually implicit high level relations (e.g.~between paragraphs), lack of hierarchical structure and nuclearity (recognizing that documents consist of more or less important but coherent parts and subparts), and a focus on a very limited inventory of signals (the latter is addressed by building on the RST-SC inventory instead).

\section{Data}\label{sec:data}

To confront our formalism with real data and provide testing grounds for linguistic and computational research using the theory, we extend the RST annotations in the English Georgetown University Multilayer corpus (GUM, \citealt{Zeldes2017}). GUM is a growing corpus, created through a classroom annotation project in which students learn to annotate a text across a semester of coursework using multiple formalisms, resulting in a rich set of annotations for each document. The corpus covers morphosyntactic annotations according to Universal Dependencies guidelines \cite{MarneffeEtAl2021}, nested entity annotations, coreference and bridging anaphora (see \citealt{Zeldes2022}), complete RST trees and more. At the time of this paper's submission, GUM (v9) encompassed $213$ documents, which come from 12 different spoken and written genres, detailed in Table \labelcref{tab:gum}. These form the data analyzed in this paper; however we note that since that time, the GUM corpus has grown to encompass four more genres in version 10 (courtroom transcripts, essays, letters and podcasts), and has been expanded with a test partition-only corpus called GENTLE (\textbf{GEN}re \textbf{T}ests for \textbf{L}inguistic \textbf{E}valuation, \citealt{aoyama-etal-2023-gentle}) with eight challenging genres.\footnote{Specifically: dictionary entries, live eSports commentary, legal, medical, poetry, mathematical proofs, syllabi and threat letters.} These data sources have at the time of publication also been annotated in the \theory{} framework, bringing the total available data up to 246K tokens in 24 genres. To browse these analyses, see the eRST website.\footnote{\url{https://gucorpling.org/erst/}}

\begin{table}[htbp]
\centering
\resizebox{0.5\columnwidth}{!}{%
\begin{tabular}{@{}llccr@{}}
\toprule
\textbf{Genres} & \textbf{Source} & \textbf{Docs} & \textbf{Tokens} & \textbf{EDUs} \\ 
\midrule
Interviews & Wikinews & 19 & 18,190 & 2,410 \\
News stories & Wikinews & 23 & 16,145 & 1,779 \\
Travel guides & Wikivoyage & 18 & 16,514 & 1,792 \\
How-to guides & wikiHow & 19 & 17,081 & 2,395 \\
Academic & various & 18 & 17,169 & 1,981 \\
Biographies & Wikipedia & 20 & 18,213 & 2,071 \\
Fiction & various & 19 & 17,510 & 2,474 \\
Web forums & Reddit & 18 & 16,364 & 2,263 \\
Conversations & SBC & 14 & 16,416 & 2,810 \\
Speeches & various & 15 & 16,720 & 1,914 \\
Vlogs & YouTube & 15 & 16,864 & 2,436 \\
Textbooks & OpenStax & 15 & 16,693 & 2,027 \\
\midrule
\multicolumn{2}{l}{total} & \textbf{213} & \textbf{203,879} & \textbf{26,352} \\
\end{tabular}%
}
\caption{Genres and Documents in the GUM Corpus, version 9.} \label{tab:gum} 
\vspace{-6pt}
\end{table}

With over $26$K EDUs in v9 (or $32$K in the recently released v10 + GENTLE), GUM is the largest English RST corpus, followed by RST-DT, with $21,789$ EDUs. GUM data is available from the corpus website (\url{https://gucorpling.org/gum/}), with underlying text under respective licenses from each source, and annotations under a Creative Commons Attribution license (CC-BY 4.0).

Due to the centrality of the WSJ corpus in past studies of RST, for this article we also partly annotate RST-DT for \theory{} annotations, focusing on the \textbf{test-set} of 38 documents, to which we add full token-anchored annotations of discourse markers (including orphans based on the primary tree) and corresponding secondary edges according to our guidelines. Due to the licensing restrictions on RST-DT, we release these annotations separately, without the underlying text.

\subsection{Annotation Process}\label{sec:annotation}

Since primary RST trees were already available for GUM and RST-DT, \theory{} annotations were divided into three main parts: 1. identifying and associating DMs with trees; 2. adding secondary edges where DMs were left over as orphans or syntactic triggers were identified; and 3. adding non-DM signals using semi-automatic methods.

\subsubsection{DM, Orphan and Secondary Edge Annotation}

For DM identification, we preprocessed the corpus with the best-performing English connective detector trained on PDTB v3, the DisCoDisCo system \cite{GesslerEtAl2021}, winner of the DISRPT 2021 shared task on Connective Detection \cite{ZeldesEtAl2021a}. This step was undertaken to ensure high recall, high conformity with PDTB connective definitions and high consistency (it has repeatedly been shown that correcting state of the art NLP outputs outperforms purely manual annotation due to tool consistency, see \citealt{mikulov-EtAl:2022:LREC}).

Following connective detection, a script associated each predicted connective with the nearest compatible relation in the tree hierarchy, prioritizing the outgoing relation of the EDU containing it, followed by recursively searching for a larger span containing the original EDU until finding a relation compatible with the connective, based on the PDTB guidelines and the PDTB-RST relation mapping from \citet{DembergEtAl2019}. If no relation is found, then the connective is flagged as an orphan. 

\begin{table}[htbp]
\centering

\begin{tabular}{l|rrrrr}
\toprule
        & untyped & typed (total) & typed (DM) & typed (orphan) & sourced \\
\midrule
P & 89.31\% & 72.73\% & 86.39\% & 23.11\% & 46.39\%\\
R & 78.83\% & 64.18\% & 68.68\% & 33.93\% & 40.88\%\\
F & 83.74\% & 68.16\% & 76.51\% & 27.50\% & 43.46\%\\
\bottomrule

\end{tabular}
\caption{DM detection and attachment performance
Untyped = connective detection; Typed = detection + classification: DM or orphan; Sourced = detection + classification + association with the correct relation.}
\label{tab:connective-attachment-score}
\vspace{-10pt}
\end{table}

\Cref{tab:connective-attachment-score} summarizes performance for connective detection and alignment. The higher precision for all metrics (except orphans) is due to the aligner searching for a compatible \textit{outgoing} relation, but not \textit{incoming} relations. Allowing attachment to any compatible incoming relation increases recall but degrades precision substantially. The current outputs were deemed sufficient as a starting point for manual correction.

After this preprocessing, five annotators went over the entire dataset manually to correct DM identification and alignment to relations, adding secondary edges for true orphans, whose relations were not expressed in the primary tree. Manual annotation was done using rstWeb \cite{Zeldes2016}, an open source web interface for RST annotation which was extended to support signal marking by \citet{GesslerEtAl2019}, and which we extend further for this article with support for secondary edges.\footnote{Available at \url{https://gucorpling.org/rstweb/info/} under the MIT license.} The annotation process was repeated for the RST-DT test set. 

We conducted an inter-annotator agreement (IAA) study of DM identification and relation association for $36$ GUM documents and the RST-DT test set, and report mutual F1 scores. For RST-DT, three annotators double-annotated 38 documents in the test partition, amounting to about $21$K tokens. A mutual F1 score of $95$ was obtained for identifying DMs, and $88.8$ for relation association. For GUM, two annotators double-annotated 36 documents (32K tokens) and obtained an F1 score of $92.3$ for DM identification and $90$ for relation association. 

For secondary edges, a first IAA experiment on the GUM dev set ($24$ documents) showed substantial disagreements, with S/N/R scores of .$311$, .$279$, and .$223$, corresponding to \% agreement on edge attachment points (regardless of source/target), exact edge path (including directionality) and the assigned relation. This is despite the fact that annotators achieved an F-score of .$642$ on detecting orphan DMs (i.e.~agreeing that a secondary edge was called for, and where the DM tokens were). Inspection revealed that disagreements hinged either on whether an orphan candidate was a connective (especially for spoken sentence initial `And' or `So') and what the exact scope of the relation included (e.g.~including or omitting trailing bibliographical citations in academic data). After refining the guidelines to be more explicit, a second experiment on an additional $12$ documents produced much better results of S/N/R = $.529$, $.49$, and $.412$ respectively, indicating agreement levels just $16$ points below the human agreement score on primary relation R of $.571$ \cite{morey-etal-2017-much}. We consider this to be substantial agreement, especially given that secondary relations involve some of the most challenging ones, and their scores do not benefit from common, easy cases such as relative or other adnominal clause attachments.

Finally, with the entire corpus in hand, one of the authors of the article reviewed all annotations for a final consistency pass and finalized the list of possible DMs, which for RST-DT was a subset of the PDTB connectives. The list of DMs in GUM required some expansions due to items that are not attested in PDTB or related corpora, such as TED-MDB, probably due to the different genres in the corpus.\footnote{Although TED-MDB contains spoken data, it does not cover dialogue, nor data from the web, such as GUM's Reddit data. Such user-generated content often contains unique words or spellings, such as `/' or `cuz' for `or' and `because', and may require adaptation of guidelines, see \citet{sanguinetti-etal-2020-treebanking,SanguinettiEtAl2022}.} Table \labelcref{tab:GUM-dm-not-in-PDTB} provides the added items, along with the rationale for adding them by analogy to existing PDTB items.

\begin{table}[h!tb]
\centering
\resizebox{\textwidth}{!}{%
\begin{tabular}{lll}
\toprule
\textbf{type} & \textbf{marker} & \textbf{by analogy to} \\
\midrule
subordinating & cuz & because \\
 & cause & because \\
 & whilst & while \\
 & as far as & as long as \\
 & into & marks \textsc{result}, e.g.~`trick someone \textbf{into} thinking' \\
 & that & marks \textsc{purpose}, e.g.~`I longed for nets, \textbf{that} I might capture them' \\
 & whither & where \\
 & wherever & whenever \\
adverbial & as such & marks \textsc{result} \\
 & apart from & aside from \\
coordinating & or else & not attested in PDTB; used for disjunction, analogous to `otherwise' \\
 & / & standing for `or' \\
prepositional & by the end & in the end \\
 & in essence & in short \\
 & at the time & at the same time \\
 & around the same time & at the same time \\
 & to that end & to this end \\
 & to wit & not in PDTB, similar to `for example' or `specifically' \\
 & in brief & in short \\
 & since then & since (used adverbially in isolation) \\
 \bottomrule
\end{tabular}
}
\caption{Twenty additional GUM DMs not attested in PDTB as connectives.} \label{tab:GUM-dm-not-in-PDTB} 
\vspace{-6pt}
\end{table}

We note that `aside from' is attested in PDTB only once as an AltLex governing a noun phrase, but in GUM we have `apart from thinking', which fits the DM definition by governing a VP. The item `as such' is attested in PDTB as AltLex as well, but should be an explicit connective since it is a relational prepositional phrase, similar to `at the same time'. All other DMs in GUM beyond the $20$ in Table \labelcref{{tab:GUM-dm-not-in-PDTB}} are attested in PDTB, amounting to a total of $211$ distinct types, disregarding connective modifier variants such as `\textit{two/three months} later', which we consider to belong to the type `later'.

\subsubsection{Non-DM Signals}

Due to limited resources, we did not annotate or correct all non-DM signals fully manually, and did not annotate them in the RST-DT data. However, thanks to the rich annotations available in GUM, we were able to induce many signal types fully automatically from the gold syntax trees and coreference annotations, and were able to manually annotate or correct many other cases, and evaluated accuracy and agreement manually on a subset of documents (see below).

\textit{Graphical} and \textit{reference} signals were tagged automatically based on token forms (parentheses, question marks, etc.) and gold coreference chains for eligible relations. For example, any \textsc{question} relation whose sentence contained a question mark was assumed to be signaled by that question mark, and any \textsc{elaboration} relation containing pronominal anaphora in a satellite pointing to an antecedent in the nucleus was automatically admitted as a \textit{reference} signal. The list of relations eligible for each such signal type was obtained by consulting the RST-SC corpus.

Some \textit{morphological} and \textit{syntactic} signals were also tagged automatically using the Python dependency tree-editing library DepEdit \cite{peng-zeldes-2018-roads}: Relative or infinitival clauses conveying adnominal attribute relations are easy to identify, as are reported speech for \textsc{attribution}, imperative mood for \textsc{motivation}, modals in a \textsc{condition}, and some tense markers (e.g.~past perfect marking \textsc{background} relations). Harder cases required manual verification, such as change of tense to signal a \textsc{sequence} (e.g.~past followed by present or present followed by future, fully verified manually), parallel syntactic constructions (annotated completely manually), or semantic attribution sources. For the latter, the subject or external subject (for nested or coordinate verb phrases) of each attribution predicate was identified using (enhanced) UD trees, and remaining cases for which the source could not be identified were annotated manually, such that every \textsc{attribution} in the corpus has a source signal.

Among the trickiest categories to annotate were \textit{lexical} signals, which require a large inventory of candidate items, and \textit{semantic} lexical chains, which consist of related, non-co-referring word or phrase pairs that signal a relation, and are open-ended. For the former, we took the combined list of AltLex expressions in PDTB, a manually filtered list of the top $100$ items most associated with each relation by chi square statistics, and additional items that were noticed during the annotation, all restricted to relevant POS tags. We observe that automatic annotation of all such items as signals for their associated relation was close to error free: This is intuitively not very surprising, since, if a word associated with \textsc{evaluation}, such as `good', appears in such an already manually annotated gold standard relation, it is highly likely to be signaling that relation. A total of only $10$ errors contradicting this approach were noticed during quality controls. To illustrate the process, consider example \labelcref{ex:non-indicative}, from a short story about a boy leading his developmentally disabled sister, Cara, out of a store after an unpleasant incident.

\ex. I herd Cara towards the front of the store, mouthing sorry at the front cashier. [She’s kind of \textbf{pretty}.]$_{<evaluation-comment>}$ She smiles at me. [\textbf{Nice} big brother with retarded sister.]$_{<evaluation-comment>}$\label{ex:non-indicative}

The first \textsc{evaluation} contains two words listed as evaluative: `kind' and `pretty'; however the instance of `kind' is in the wrong part of speech (`kind' is listed as evaluative only as an adjective), so only `pretty' is selected as a signal. In the second \textsc{evaluation} about the cashier's smile, `Nice' is correctly identified; a false positive, adjectival `big' would also be included as a signal, and constitutes one of the $10$ errors noticed in the data and removed manually.

For lexical chains, we were concerned about creating a subjective list of associated word pairs from the corpus, and instead decided to use a large existing inventory of word associations, opting for MIT's ConceptNet \cite{SpeerEtAl2017}, which contains over $34$ million conceptual relations between words. We allowed a script to suggest lexical chains of two or more items in the sentence containing each connected satellite and nucleus, or in two connected clauses for intra-sentential relations, and filtered the result manually. Since ConceptNet does not connect items to themselves, but lexical repetitions or variations on the same stem (e.g.~`participate'...`participant') can be signals, especially for \textsc{restatement} and \textsc{preparation} relations, we also allowed candidates based on stem matching using the Snowball stemmer for English \cite{Porter2001}. Examples of both types of chain appear in \labelcref{ex:lexchain1}--\labelcref{ex:lexchain2}:

\ex. He had no political \textbf{power}, [and his \textbf{influence} extended only so far as he was humored by those around him]$_{<elaboration-additional>}$\label{ex:lexchain1}

\ex. [Have a realistic but \textbf{exaggerated} setup.]$_{<organization-preparation>}$ The opening of the joke — or setup — should have a basis in the real world so your audience can relate to it, but it should also include \textbf{exaggeration} \label{ex:lexchain2}

In \labelcref{ex:lexchain1} `influence' is recognized as a type of `power', mirrored by a ConceptNet `is-a' relation between the two words, while in \labelcref{ex:lexchain2}, the identical stem `exaggerat-' helps to identify how the initial \textsc{preparation} satellite prepares the reader for the subsequent nucleus.

The final list of lexical chains amounted to $1,280$ manually verified instances, covering $2,825$ tokens in the corpus, meaning approximately $1.3$\% of corpus tokens are part of a lexical chain. We note that the total would have been much higher due to plain repetition of nouns, e.g.~in \textsc{elaboration} or \textsc{background} relations, but many of these were rejected from the chain proposing script, not because they were irrelevant, but because they were already captured under coreference-related signal types, and were therefore ineligible to be lexical chains (e.g.~the `setup', mentioned twice in \labelcref{ex:lexchain2}, is excluded as a lexical chain, because it is already part of a coreference-based signal instead).

To evaluate the accuracy of our annotations, and humans annotators' ability to agree on non-DM signals, we conducted two experiments: annotations from all signal categories were manually corrected for $12$ documents in the test set (one from each genre), and four of these were double annotated. Table \labelcref{tab:iaa-non-dm} gives mutual precision/recall and F1-scores for the humans (hum-v-hum, $4$ docs) and for the automatic annotation performance compared to the human annotation (cpu-v-hum). We evaluate in two scenarios: token-anchored (signals only match if their type, subtype, covered tokens and associated relation match) and unanchored (token spans may differ), and we report micro and macro-averaged scores (across documents average).

\begin{table}[h!tb]
\centering
\small
\begin{tabular}{l|lll|lll}
\toprule
                   & anchored &       &       & \multicolumn{2}{l}{unanchored} &       \\
                   \midrule
hum-v-hum          & P        & R     & F1    & P              & R             & F1    \\
\midrule
\textit{Micro}     & 0.813    & 0.798 & 0.805 & 0.865          & 0.837         & 0.851 \\
\textit{Macro}     & 0.809    & 0.801 & 0.804 & 0.859          & 0.839         & 0.848 \\
\midrule
\textit{cpu-v-hum} & P        & R     & F1    & P              & R             & F1    \\
\midrule
Micro              & 0.841    & 0.92  & 0.879 & 0.868          & 0.95          & 0.907 \\
Macro              & 0.837    & 0.922 & 0.877 & 0.865          & 0.954         & 0.906 \\
\bottomrule
\end{tabular}
\caption{Human vs. human and system vs. human agreement for all signal types on a subset of documents.} \label{tab:iaa-non-dm} 
\vspace{-8pt}
\end{table}

Comparing human vs. human scores (top half of the table) to automatic system scores (bottom) is not strictly possible, since human agreement is computed on a smaller subset of documents; that said, we can observe that the system performs about as well as humans (on precision) or better (on recall), and that the gap between anchored and unanchored scenarios is similar as well. We suspect that the reason why the system has the upper hand in recall is that two humans inevitably generate additional disagreeing signals, while they are less likely to remove a predicted signal unless it is truly wrong. As a result, when compared with adjudicated output containing only the more unanimously recognized signals, the system does not miss less certain cases which one human might flag but not another. 

Looking at prediction errors more qualitatively, we note that lexical chain disagreements are by far the most common in both human and system performance, followed by indicative words, while syntactic and coreference based signals are almost always correct. To understand why, we consider the typical example type in \labelcref{ex:lexchain-err}--\labelcref{ex:lexchain-err2}, compared to the very uncommon syntactic error type in \labelcref{ex:syn-err}.

\ex. The Beavertail \textbf{cactus} can grow to be about 24 inches ... has \textbf{pads} that look like beavertails\label{ex:lexchain-err}

\ex. There are \textbf{holes} in the center of the base of each pot to allow for \textbf{drainage}\label{ex:lexchain-err2}

\ex. A: Do you need a partner? B: \textbf{To} go there?\label{ex:syn-err}

In \labelcref{ex:lexchain-err}, an annotator recognizes `cactus' and `pads' as a meaningful lexical chain for a \textsc{list} relation, a pair of related terms since pads are the leaves of a cactus; however the second annotator does not recognize this chain, and they are not included as related terms in ConceptNet. In \labelcref{ex:lexchain-err2}, one annotator recognizes `holes' and `drainage' as a chain indicating a \textsc{purpose} relation, which is again not detected by another annotator or ConceptNet. Finally in \labelcref{ex:syn-err}, we see an unusual case of a syntactic signal across speakers, where a human annotator marks the infinitive `to' as a syntactic signal indicating \textsc{purpose}, but the system misses the signal, since these are separate sentences and there is therefore no syntactic tree relation to indicate the signals -- such examples of syntactic signal corrections are very rare. Overall, we feel these results indicate a high level of reliability for the additional signal types in the corpus, while also indicating that subtypes such as lexical chains and other indicative lexical signals warrant more study in order to arrive at dependable operationalizations that do not rely solely on a lexical resource like ConceptNet.

\subsection{\theory{} Annotations across Genres}

Table \labelcref{tab:data-dm-secedge} gives an overview of the prevalence of secondary edges in GUM as a whole, as well as by genre, and in comparison to our annotations of the RST-DT test-set. As the table shows, $\sim$$13$\% of discourse markers in GUM are orphans ($895$/$6,920$), fewer than in RST-DT, which has $\sim$$17$\% ($87$/$493$). At the same time, the proportion of secondary edges is identical in both datasets at $3.37$\%, and there are only slightly more DMs per relation in GUM (.$231$ on average per relation, compared to .$191$ in RST-DT). The latter differences suggest different amounts of unmarked relations in both corpora, and slightly more relations with multiple markers in GUM.\footnote{An anonymous reviewer has noted <$4$\% of relations being secondary may mean that they are close to negligible, but we note that depending on our interests they can be very important: over $14$\% of GUM \textsc{result} relations are secondary, as are over $10$\% of \textsc{concession} relations (see Section \labelcref{sec:applications} for more  statistics). These are highly relevant to semantic applications, research on political speech, and more, and demonstrate the added value of \theory{} in covering the full breadth of relations in texts.}

\begin{table}[b!htp]
\centering
\resizebox{\textwidth}{!}{%
\begin{tabular}{l|rrrrr|rr}
\toprule
        genre &   dms &  orphans &  dms+orphans &  relations & markers\_per\_rel &  secedges &  \%secedge  \\
\midrule
RST-DT & 406 & 87 & 493 & 2,580 & 0.191 & 87 & 3.37\% \\
\midrule
GUM & 6,025 & 895 & 6,920 & 29,950 & 0.231 & 1,008 & 3.37\% \\
\midrule
news & 334 & 38 & 372 & 1,933 & 0.192 & 43 & 2.22\% \\
academic & 403 & 55 & 458 & 2,069 & 0.221 & 61 & 2.95\% \\
bio & 392 & 38 & 430 & 2,303 & 0.186 & 53 & 2.30\% \\
conversation & 484 & 127 & 611 & 3,341 & 0.182 & 131 & 3.92\% \\
fiction & 611 & 85 & 696 & 2,899 & 0.240 & 97 & 3.35\% \\
interview & 514 & 73 & 587 & 2,698 & 0.217 & 83 & 3.08\% \\
reddit & 622 & 89 & 711 & 2,606 & 0.272 & 100 & 3.84\% \\
speech & 446 & 70 & 516 & 2,179 & 0.236 & 76 & 3.49\% \\
textbook & 440 & 47 & 487 & 2,201 & 0.221 & 54 & 2.45\% \\
vlog & 675 & 181 & 856 & 2,942 & 0.290 & 193 & 6.56\% \\
voyage & 397 & 50 & 447 & 2,062 & 0.216 & 64 & 3.10\% \\
whow & 707 & 42 & 749 & 2,717 & 0.275 & 53 & 1.95\% \\

\bottomrule

\end{tabular}
}
\caption{DMs and Secondary Edges in RST-DT test, GUM, and the GUM genres.} \label{tab:data-dm-secedge} 
\vspace{-8pt}
\end{table}

Looking at GUM genres, we see considerable variation. \textit{News} unsurprisingly comes very close to the RST-DT values for DMs, but far below for secondary edges, perhaps because GUM news stories are shorter than some of the long and complex texts in the RST-DT test set. Other genres are even stronger outliers: Secondary edges are rarest in how-to guides (\textit{whow}, $1.95$\%) and most common in \textit{vlog} ($6.56$\%), the latter due to frequent linking of sentences with sentence initial \textit{And}, and to some extent \textit{So} (only counting cases in which `so' is actually a DM). \textit{Academic} is surprisingly below average in DMs per relation, despite common assumptions in the literature about the explicitness of academic text (e.g.~\citealt{Hughes1996}; see \citealt{BiberGray2010} for criticism).

\begin{figure}[h!tb]
    \centering
    \includegraphics[width=0.95\textwidth]{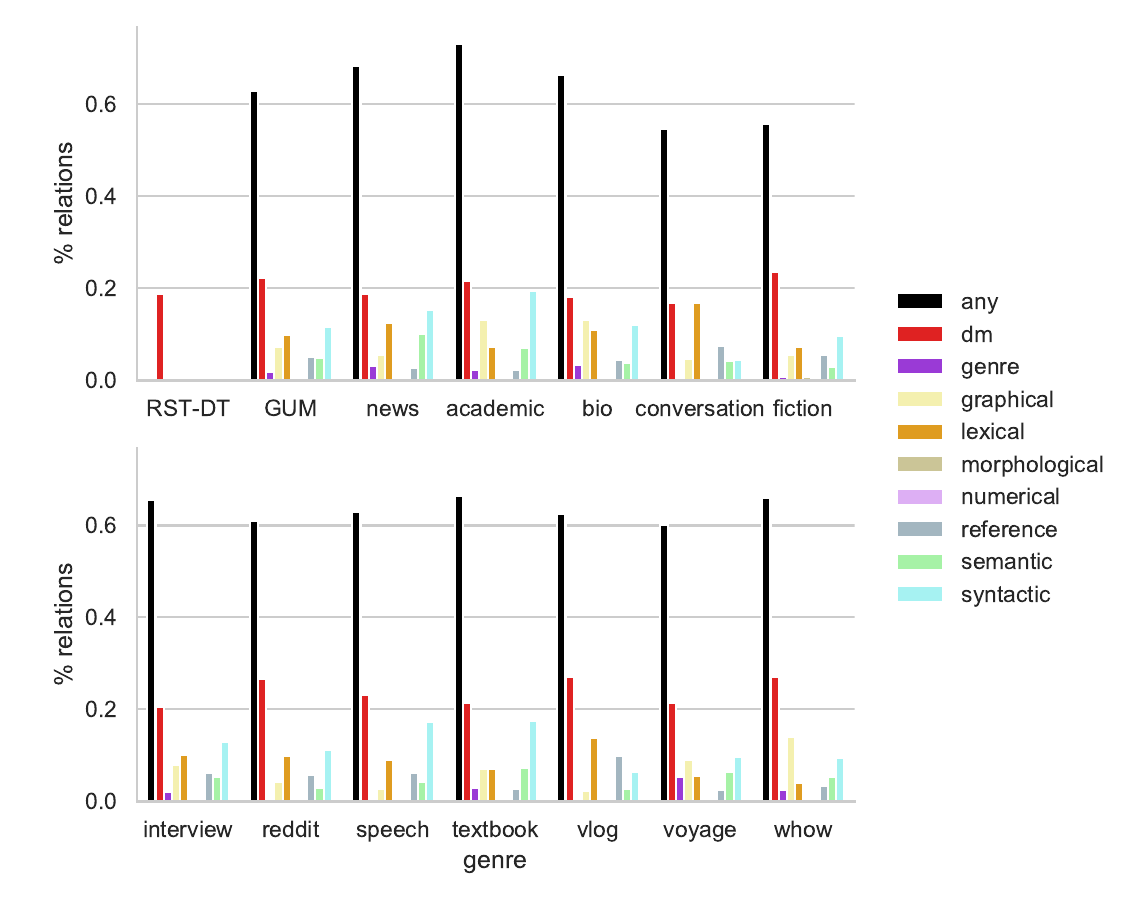}
    \vspace{-4pt}
    \caption{
    Distribution of signal types across genres vs. GUM and the RST-DT test set.}
    \label{fig:sigtype-barplots}
    \vspace{-12pt}
\end{figure}

However, it would be wrong to say that \textit{vlogs} are more explicit in their discourse relations than \textit{news}, since DMs/orphans are not the only means of signaling relations. Turning to all signal types, Figure \labelcref{fig:sigtype-barplots} shows the percentage of each major signal class across all GUM genres vs. the entire corpus, including the proportion of relations signaled by any means in black (`any'). RST-DT is shown for comparison but only has data for DMs (the `dm' bar includes orphans, and secondary edges are counted in relation totals).

As we can see GUM \textit{news} is very comparable to RST-DT in DM prevalence. We also observe that \textit{academic} is not actually low in signals: It marks the most relations at $73.2$\%, followed by \textit{news} at $68.4$\%, \textit{textbook} ($66.4$\%) and \textit{whow} ($66.1$\%), mainly owing to syntactic cues in the first three, and to some extent frequent graphical signals as well (esp. \textit{whow}). The overall `any' signal proportion in GUM is $63$\%, much lower than the figure by \citet{DasTaboada2017} for RST-DT at $92.7$\%, though we note that some different signal types were included in that study, and that GUM \textit{news} is closer to that mark.\footnote{The lack of token-level searchable annotations in RST-SC complicates tracking down causes for these differences quantitatively, but data inspection suggests the inclusion of much more wide ranging lexical chains constitutes most of the difference.} By contrast, the finding of around $20$\% DM marking in RST-DT test is in line with a similar estimate of $18.21$\% in \citet[26]{DasTaboada2017}.

\begin{figure}[h!tb]
    \centering
    \includegraphics[width=1\textwidth]{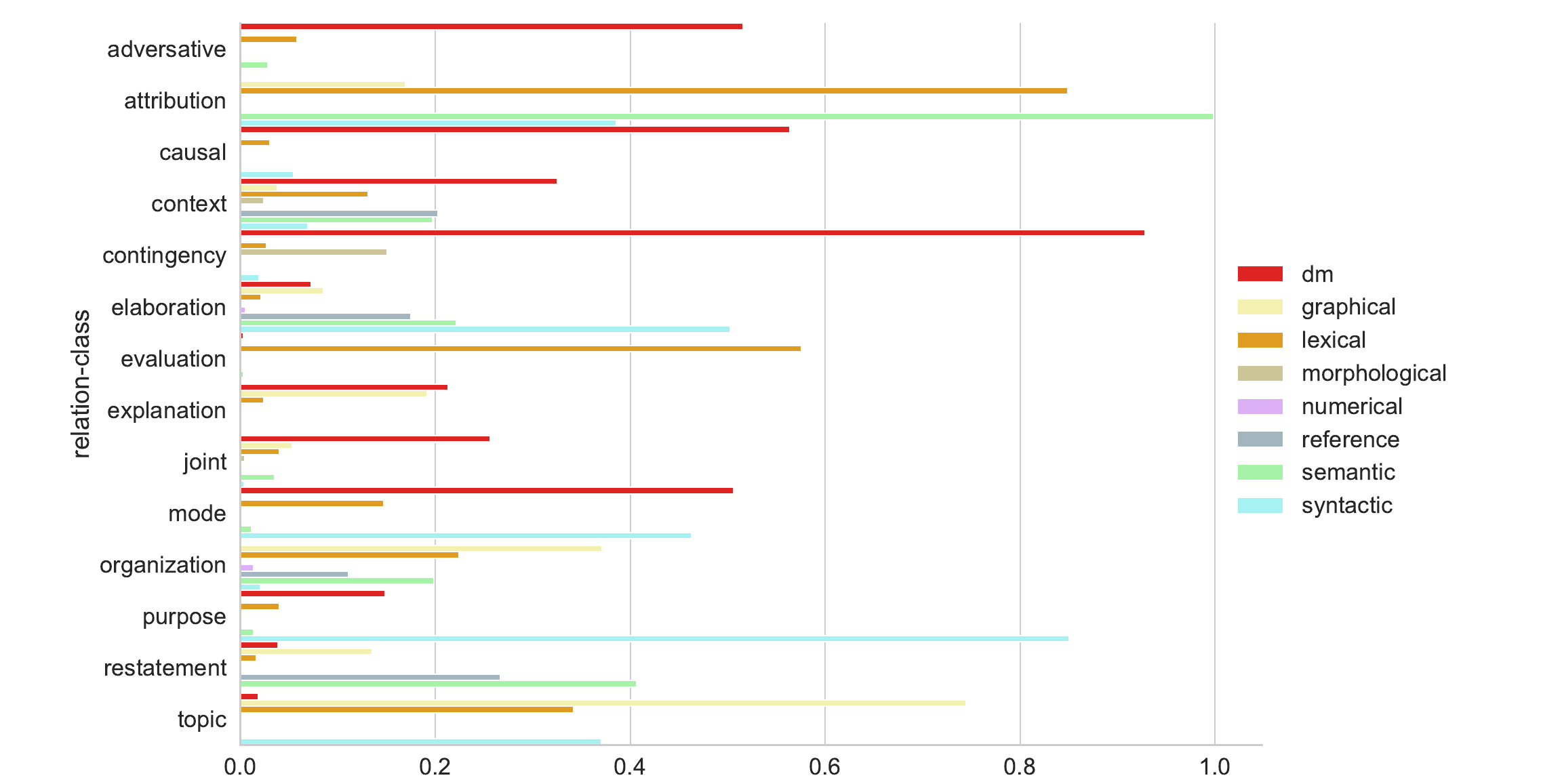}
    \caption{
    Distribution of Signal Types for Each Coarse Relation Class in GUM.}
    \label{fig:sigrel-barplots}
    \vspace{-4pt}
    
\end{figure}

Another set of contrasts obscured by looking just at signal types can be seen by comparing how each coarse relation class tends to be signaled, which is very heterogeneous.\footnote{The same can be said for fine-grained relations, which we disregard here for space reasons.} Figure \labelcref{fig:sigrel-barplots} gives the proportion of signals for each relation class. We can see some relations are almost never signaled by a DM (for example \textsc{attribution}, \textsc{organization}), while others nearly always have one (\textsc{contingency}, normally marked by \textit{if}, but sometimes by other means, such as syntactic inversion or morphological mood), and combinations are quite common (cf.~\citealt{Crible2022}). The \textsc{purpose} class is marked almost always by a syntactic signal (usually a \textit{to}-infinitive), but can have other signals, e.g.~the DM \textit{in order}). Relations which can be signaled by recurring mentions, such as \textsc{elboration}, \textsc{restatement} and \textsc{context}, show such \textit{semantic} and \textit{reference} signals often (especially \textit{semantic repetition}). We can also see that \textsc{attribution} is always marked (not least because a \textit{semantic source} for the attribution should be present by definition), while the least marked class overall is \textsc{joint}, containing e.g.~temporal \textsc{sequence} (often marked only by implicit chronological order), \textsc{list}, and other coordinate structures.

The most lexically marked classes are \textsc{evaluation} (e.g.~positive and negative adjectives) and \textsc{organization}, the latter primarily due to back-channeling and preparation markers in conversation (`Uh-huh', `you know,', `I mean,'). Since our data delivers not only statistics on relation signaling types, but actual aligned token indices for each relation's signals, we can extract the most frequent DMs and lexical expressions used to convey each relation, which are given in Table \labelcref{tab:top-dm-lex}.

\begin{table}[h!tb]
\centering
\resizebox{0.95\textwidth}{!}{%
\begin{tabular}{l|rlll}
\toprule
relation & freq & \%signaled & top DMs & top lexical \\
\midrule
adversative & 2,405 & 55.88\% & but (641), however (101), though (59), and (59) & may (35), only (24), might (14), actually (13) \\
attribution & 1,592 & 99.94\% & as (2) & said (162), think (152), know (113), say (73) \\
causal & 1,240 & 63.31\% & and (221), because (167), so (167), as (22) & due to (7), result (6), caused (6), cause (3) \\
context & 2,317 & 79.59\% & when (311), as (96), after (81), while (48) & never (17), always (17), often (12), following (11) \\
contingency & 518 & 96.72\% & if (367), if then (35), when (31), unless (12) & depending (4), based on (3), every time (2), in the case (2) \\
elaboration & 5,339 & 89.32\% & and (208), also (55), for example (28), with (25) & including (36), too (18), e.g. (17), especially (14) \\
evaluation & 1,047 & 57.21\% & and (5), so (4), so that (1) & good (45), very (40), important (19), great (19) \\
explanation & 1,650 & 41.94\% & so (104), because (69), as (42), and (22) & see (22), shown (5), based on (3), considering (3) \\
joint & 8,922 & 34.17\% & and (1701), also (211), then (202), or (127) & now (39), too (34), again (24), today (12) \\
mode & 512 & 78.71\% & by (98), as (76), without (32), as if (14) & using (44), based on (19), according to (5), guided by (2) \\
organization & 1,805 & 75.73\% & thus (1) & you know (94), yeah (57), oh (54), I mean (42) \\
purpose & 904 & 94.03\% & for (39), so (30), in order (24), so that (18) & stop (7), achieve (6), prevent (5), avoid (4) \\
restatement & 1,213 & 55.56\% & and (23), in other words (7), or (7), in short (2) & that is (6), aka (6), i.e. (5), known as (2) \\
topic & 486 & 82.51\% & so (6), if (3) & what (80), how (42), why (24), who (11) \\

\bottomrule
\end{tabular}

}
\caption{Top DMs and lexical markers per coarse relation class.} \label{tab:top-dm-lex} 
\vspace{-6pt}

\end{table}

As the table shows, some of the most frequent DMs are unsurprisingly polysemous (`and', `so' and `as' occur in several classes), and next to DMs we find lexical signals for every class, which often work in tandem with the DMs, e.g.~the DM `then' in the \textsc{joint} class can co-occur with lexical items such as `today' to mark temporal sequences, or with adverbs not recognized as DMs by PDTB, such as `too' for a \textsc{joint-list}. And while \textsc{attribution} is only marked by a DM twice, as in \labelcref{ex:attrib-as}, it is usually accompanied by a speech or cognition verb such as `said' or `think', providing a clear lexical signal.

\ex. $[$\textbf{As} Heald told The Huffington Post,$]_{<attribution>}$ US surface ozone has dropped partly due to the Clean Air Act.\label{ex:attrib-as}

The data in the table only begins to scratch the surface of how relations are marked, and much more remains to be learned by examining the ways in which relations can conceivably be marked, and ways in which the same items may occur without signaling the relation with which they are associated, a topic we leave to future research.

\subsection{Search and Visualization}

To support exploration of the data, we add support for \theory{} annotations to the existing RST search and visualization facilities in ANNIS (\citealt{KrauseZeldes2016}, available open source under the Apache 2.0 license), an open source search platform for multilayer corpus data.\footnote{Available from \url{https://corpus-tools.org/annis/}} ANNIS supports search across all annotation layers in GUM, meaning queries can combine syntactic structures, coreference links, RST relations and more. For instance, the ANNIS Query Language (AQL) example in \labelcref{ex:annis1} searches for a non-terminal \texttt{group} (a complex discourse unit) dominating a terminal \texttt{segment} (an EDU) with some \textsc{adversative} relation type, using a regular expression, with a representative result shown below it (node colors in the query correspond to covered text color in the example).

\ex. \small \texttt{\textcolor{red}{kind}="group" >[relname=/adversative.*/] \textcolor{magenta}{kind}="segment"}\label{ex:annis1} 
\a. \textit{$[$\textcolor{red}{The value of Airbnb is approximately \$30 billion.}$]$ $[$ \textcolor{magenta}{Compare this market value to Hilton ’s market capitalization of \$19 billion and Marriott’s of \$35 billion.}$]$$_{<adversative-contrast>}$}
  \b. \textit{$[$\textcolor{magenta}{Not with your gloves or anything .}$]$$_{<adversative-antithesis>}$ $[$ \textcolor{red}{Find something else to pick it up with.}$]$}

Using the expansion to the functionality to support \theory{}, we can also limit results to ones in which a DM signal is available, and anchored to a word with a particular POS tag, for example a coordinating conjunction (POS tag \texttt{CC}), marked in green in \labelcref{ex:annis2}. The operator \texttt{\_i\_} indicates that the second node \textit{includes} the POS node, and the expression \texttt{\textcolor{olive}{signal\_type}="dm" > \textcolor{ForestGreen}{\#3}} indicates that a DM signal dominates the token declared as the third node, which carries the POS tag \texttt{CC}.

\ex. \small \texttt{\textcolor{red}{kind}="segment" >[relname=/adversative.*/] \textcolor{magenta}{kind}="segment"  \_i\_ \textcolor{ForestGreen}{pos}="CC" \& \textcolor{olive}{signal\_type}="dm" > \textcolor{ForestGreen}{\#3}}\label{ex:annis2}
\a. \textit{$[$\textcolor{red}{They have ideas}$]$ $[$\textcolor{ForestGreen}{but} \textcolor{magenta}{they can not formulate them in the right way.}$]$$_{<adversative-contrast>}$}
 \b. \textit{$[$\textcolor{red}{We should strive for equality}$]$ $[$\textcolor{ForestGreen}{but} \textcolor{magenta}{also practicality.}$]$$_{<adversative-contrast>}$}

\begin{figure}[h!tb]

  \includegraphics[clip,width=0.4\textwidth,trim = 0cm 0cm 0cm 0cm]{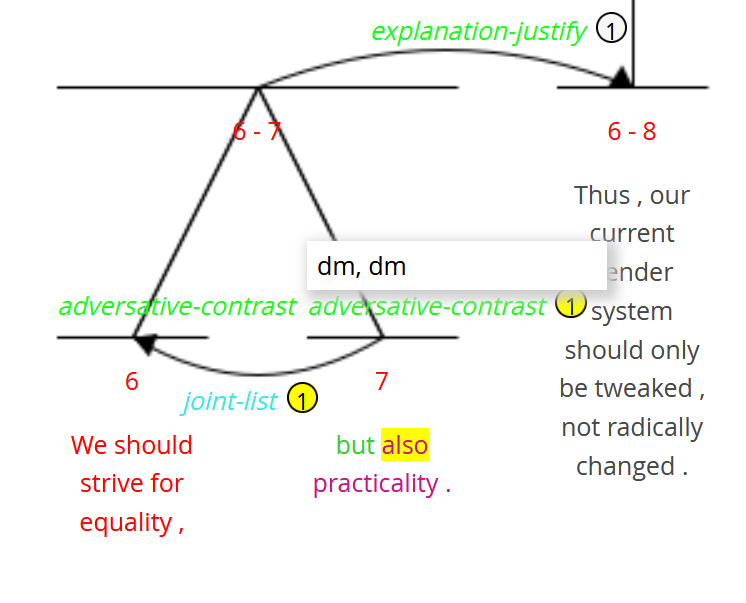}%
\centering
\vspace{-6pt}
\caption{ANNIS visualization for the second result of the query in \labelcref{ex:annis2}. The visualization closely follows the view in rstWeb, except for editing facilities.}\label{fig:annis}
\vspace{-10pt}
\end{figure}

In the last example, where the \texttt{CC} is matched by the word `but', we also see a second marker `also' which is not highlighted in the query result - this corresponds to a secondary edge orphan DM, whose edge can be seen in the ANNIS visualization for this search result in Figure \labelcref{fig:annis}. Each signal can be highlighted by hovering over the counter button next to each relation (showing `1' for the \textsc{contrast} relation). The secondary edge corresponding to `also' has the relation \textsc{joint-list} connecting [6] and [7]. We release the code for the new visualization and search capabilities, and make the annotated corpus freely available for search via ANNIS at \url{https://gucorpling.org/annis}.

\vspace{-10pt}

\section{Parsing \theory{}}\label{sec:parsing}

\subsection{Task Definition and Metrics}

Using notation from Section \labelcref{sec:primary-tree}, the goal of conventional RST parsing is to produce the tree $G$ given the textual tokens $T$ and the EDUs $S$, which partition $T$ into contiguous spans.%
\footnote{Some parsers relax the requirement that $S$ be given at prediction time, but most assume $S$ as an input---i.e., the parser receives gold EDUs and segmentation is considered a separate task.}
RST parsing is usually evaluated with the Parseval metrics, and 
we follow previous work in considering only binarized trees and using the original Parseval scoring scheme instead of the older RSTParseval \citep{black-etal-1991-procedure,morey-etal-2017-much}. 
Each non-terminal vertex can be seen as the product of a \textit{parsing decision}, where two vertices $\{v^a, v^b\}$ are joined by a relation with nuclearity $n$ and label $l$.
We refer to the unordered pair $\{v^a, v^b\}$ as the decision's associated span $s$.
For any well-formed sequence of parsing decisions $D = \langle \langle s_1, n_1, l_1 \rangle, \ldots, \langle s_m, n_{m}, l_{m} \rangle \rangle$, there is exactly one tree that may result, so evaluating decisions is equivalent to evaluating the tree.

To evaluate parser output, consider the gold parsing decision sequence $D$, and the parser's decision sequence $\hat{D}$, and let $d.s$, $d.n$, and $d.l$ correspond to span, nuclearity, and label of a single parsing decision $d$.
The four Parseval metrics can now be defined as precision\footnote{We may add metrics for recall/F1, but these would only differ from precision if the assumption of tree projectivity were dropped. 
}    
 metrics over the sets $D$ and $\hat{D}$:
\begin{equation}
\small
\begin{aligned}
    \mathbf{S}\text{pan}(D, \hat{D}) & = \frac{\#\{\hat{d} \mid \hat{d} \in \hat{D} \land \exists d \in D[d.s = \hat{d}.s]\}}{\#\hat{D}}  \\
    \mathbf{N}\text{uclearity}(D, \hat{D}) & = \frac{\#\{\hat{d} \mid \hat{d} \in \hat{D} \land \exists d \in D[d.s = \hat{d}.s \land d.n = \hat{d}.n ]\}}{\#\hat{D}}  \\
    \mathbf{R}\text{elation}(D, \hat{D}) & = \frac{\#\{\hat{d} \mid \hat{d} \in \hat{D} \land \exists d \in D[d.s = \hat{d}.s \land d.l = \hat{d}.l ]\}}{\#\hat{D}}  \\
    \mathbf{F}\text{ull}(D, \hat{D}) & = \frac{\#\{\hat{d} \mid \hat{d} \in \hat{D} \land \exists\ d \in D[d.s = \hat{d}.s \land d.n = \hat{d}.n \land d.l = \hat{d}.l ]\}}{\#\hat{D}}  \\
\end{aligned}
\end{equation}

Note that all metrics depend first on checking whether some predicted span exists in the gold tree.
The Span metric does only this, and the remaining three metrics add criteria: Nuclearity and Relation metrics also require the span's nuclearity and relation label, respectively, to match the corresponding span in the gold tree, and the Full metric requires matching span, nuclearity, and relation.

\theory{} introduces signals and secondary edges, which must be scored as well.
First, let us extend our formalization
so that in addition to the vertices joined by the edges $v_i, v_j$, each member of $E$ also carries a binary indicator variable $\sigma$ which is true only for secondary edges. 
Further, we expand $G$ with $\Lambda$, a vocabulary of signal labels, and $Z$, the signals, where each signal may be defined as $\langle e, \lambda, \tau \rangle$: $e$ is the associated edge,\footnote{We will consider two edges equal if the terminal vertices of both the source and the target nodes are identical, i.e.~they span the same EDUs.} $\lambda$ is the signal type label of the signal,\footnote{This label may be decomposed into $\lambda_1$, $\lambda_2$, etc.
if desired; we use a single label item here for simplicity, e.g.~`semantic:lexical\_chain' can be a monolithic signal type.} and $\tau$ is a set of tokens associated with the signal.

Let us define signal precision $\mathbf{S}_\mathbf{P}$ and signal recall $\mathbf{S}_\mathbf{R}$, which assess the quality of the predicted signals without considering associated tokens:
\begin{equation}
\small
\begin{aligned}
    \mathbf{S}_\mathbf{P}(G, G^\prime) = & 
            \frac{\textsc{sum}(\{\#(\hat{\zeta} \cap \zeta) \mid 
                 \hat{\zeta} \subseteq \hat{Z} \land \zeta \subseteq Z \land \forall z \in \zeta [\forall \hat{z} \in \hat{\zeta}[\hat{z}.e = z.e \land \hat{z}.\lambda = z.\lambda ]]\})}
                 {\#\hat{Z}} \\    
    \mathbf{S}_\mathbf{R}(G, G^\prime) = & 
            \frac{\textsc{sum}(\{\#(\hat{\zeta} \cap \zeta) \mid 
                 \hat{\zeta} \subseteq \hat{Z} \land \zeta \subseteq Z \land \forall z \in \zeta [\forall \hat{z} \in \hat{\zeta}[\hat{z}.e = z.e \land \hat{z}.\lambda = z.\lambda ]]\})}
                 {\#Z} \\                     
\end{aligned}
\end{equation}
Informally, these equations group signals by the combination of their label and associated edge. 
This group is computed for both the gold and predicted tree, and the overlap between the gold and the predicted group for the label--edge combination is noted.
The size of this overlap is summed across all groups, and the sum is divided by the total number of predicted signals (precision) or gold signals (recall).
This slightly complicated formulation of precision and recall is necessitated by the fact that there could be potentially many signals which share the same edge and label (but then not the same tokens, which are however ignored in the above metric).

We would also like to have a quantitative \textit{anchored} metric of how well the parser performed at predicting the actual tokens (word forms) associated with a signal.
Due to the complication just noted above, formulating precision and recall metrics is not entirely straightforward.
Consider the case where for some given edge $e$ and signal label $\lambda$, the gold tree has some signals $\zeta \subseteq Z$ and the predicted tree has some signals $\hat{\zeta} \subseteq \hat{Z}$.
It is possible that even if all signals in $\hat{\zeta}$ and $\zeta$ are associated with edge $e$ and are labeled $\lambda$, $\zeta$ may not be equal in size to $\hat{\zeta}$. 
Moreover, even if they are the same size, it is not immediately clear how to put signals from the two sets into pairwise correspondence so that their tokens may be compared.

Our solution is to allow the evaluation procedure to find the optimal pairing for predicted and gold signals for each label--edge signal group.
While in principle this is an expensive operation with computational complexity $\mathcal{O}(Z!)$, we expect that a label--edge group in the $99$th percentile by size would contain no more than $5$ signals, and moreover, heuristics would likely be able to make the optimal pairing search more efficient as needed.
Let us therefore define an $\textsc{optimal-pair}(\lambda, e, Z, \hat{Z})$ function which makes label--edge groups from the signal sets and produces a set of optimal pairings $\{\langle z_1, \hat{z}_1 \rangle, \ldots, \langle z_m, \hat{z}_m \rangle\}$ that maximizes the total number of overlapping tokens across the two label--edge groups.
With the assistance of \textsc{optimal-pair}, we may now define signal token precision $\mathbf{W}_\mathbf{P}$ and recall $\mathbf{W}_\mathbf{R}$ as follows:

\vspace{-4pt}
\begin{equation}
\small
\begin{aligned}
    \mathbf{W}_\mathbf{P}(G, G^\prime) = & 
        \frac{\textsc{sum}(\{\#(\hat{z}.\tau \cap z.\tau) \mid \lambda \in \hat{\Lambda} \land e \in \hat{E} \land \langle z, \hat{z} \rangle \in \textsc{optimal-pair}(\lambda, e, Z, \hat{Z}) \})}
             {\textsc{sum}(\{\#\hat{z}.\tau \mid \hat{z} \in \hat{Z}\})} \\
    \mathbf{W}_\mathbf{R}(G, G^\prime) = & 
        \frac{\textsc{sum}(\{\#(\hat{z}.\tau \cap z.\tau) \mid \lambda \in \Lambda \land e \in E \land \langle z, \hat{z} \rangle \in \textsc{optimal-pair}(\lambda, e, Z, \hat{Z}) \})}
             {\textsc{sum}(\{\#z.\tau \mid z \in Z\})} \\             
\end{aligned}
\end{equation}
Intuitively, think of these metrics as proceeding in the following way: For every relation in the predicted and gold trees, group signals such that each group shares some edge $e$ and some signal label $\lambda$.
(Perhaps, for example, the signals all share the label $\textsc{dm}$.)
Note that (unlabeled) edge correctness is a prerequisite for signals to be deemed correct, since associated edges are identified by the EDU yield of their source and target nodes. 
Now, for each signal label--edge group, find an optimal pairing between predicted and gold signals such that overlap in each pair's associated tokens is maximized.
Finally, count the number of overlapping tokens across all pairs, and divide by the total number of token associations in either the predicted (precision) or gold (recall) signals.

For secondary edges, the four Parseval metrics are directly applicable: a secondary edge has all the salient properties of a primary edge, although we note that the term `nuclearity' should properly be replaced by `direction', since secondary edge source and target designations imply only a direction and not necessarily a higher prominence for the target vis-a-vis the source (such prominence is only represented via the primary tree, to maintain RST's unambiguous nuclearity property).
The only outstanding question for how to perform Parseval evaluation for \theory{} trees, then, is how to combine metrics that are respectively computed for the primary and secondary edges.
We expect that in general it could be useful to consider both in isolation and also to consider the metrics for both kinds of edges combined.
However, in the latter case, since secondary edges are rare, they would not change a metric pooling both very much, so in the interest of space we report on each type separately in our experiments below. We publicly release our scorer with the code and data for this paper.

\subsection{Model Architecture}

Although the main objective of this article is to present \theory{} as a framework, and providing a comprehensive NLP system for its parsing is outside of the current scope, we present an initial experiment in \theory{} parsing in this section, extending existing architectures. Conventional RST parsers take either a top-down or a bottom-up approach: Top-down begins with the entire document and decomposes it recursively into sections, which may coincide (or be forced to coincide) with paragraphs, sentences, etc. \citep{feng-hirst-2014-linear, Kobayashi_Hirao_Kamigaito_Okumura_Nagata_2020}.
Recent approaches rely on end-to-end neural architectures: The DMRST parser \citep{liu-etal-2021-dmrst} used a pointer network as its decoder and maintains a stack by top-down, depth-first span splitting; \citet{zhang-etal-2021-adversarial} employed adversarial learning to distinguish gold versus incorrect trees.

Bottom-up approaches are perhaps closer to human RST annotation practices \citep{shen-etal-2022-easy}, beginning by connecting related clauses and sentences, then larger structures. This approach currently wins on span and nuclearity identification scores, but not on relation classification. 
\citet{guz-etal-2020-unleashing} provided a transition-based neural shift-reduce parser using SpanBERT embeddings and \citet{yu-etal-2022-rst} proposed a second EDU-level pretraining on top of sentence-level training for next EDU prediction as well as discourse marker prediction. 

After briefly considering ways of implementing novel end-to-end approaches to the task above, we quickly realized that substantial additional research would be needed in order to not only add model components to predict signals and secondary edges, but also to perform at near-state of the art (SOTA) levels for primary tree parsing. \theory{} involves aspects not only of the RST parsing literature surveyed above, but also of connective detection (see \citealt{yu-etal-2019-gumdrop,GesslerEtAl2021,metheniti-etal-2023-discut}  for recent work), explicit and implicit relation recognition \cite{rutherford-etal-2017-systematic,dai-huang-2018-improving,kim-etal-2020-implicit,scholman-etal-2021-comparison}, and discourse relation classification \cite{liu-etal-2023-hits}, which remain challenging even for recent neural models \cite{qin-etal-2017-adversarial,kurfali-ostling-2021-probing,braud-etal-2023-disrpt}.  As a starting point, we therefore decided to adapt existing SOTA models for predicting primary trees and explicit connectives, and to construct a baseline approach on top of those systems.\footnote{We do not mean to say that a unified, end-to-end approach to the task is a bad idea: Our approach is merely motivated by the observation that our initial attempts to do so resulted in unusable primary tree prediction accuracy. We believe there is great potential in jointly learning the related subtasks in \theory{}, similarly to successful work in multitask learning and pretraining for RST parsing \cite{BraudPlankSoegaard2016}.} 
Our approach consists of the following four components:

\paragraph{Primary Tree Parsing} After testing several off-the-shelf parsers, we chose the top-down DMRST \cite{liu-etal-2021-dmrst}, which remains SOTA for the RST Relation metric for GUM and is efficient and easy to run. The system produces projective, binary, labeled trees, which we use as an input for signal prediction and association, as well as the basis for the available non-terminal nodes for secondary edge prediction. For comparison, we also provide numbers using the best bottom-up shift-reduce-style parser from \cite{guz-carenini-2020-coreference} in the next section.

\paragraph{Connective Detection} We use DisCoDisCo \cite{GesslerEtAl2021}, the highest scoring connective detection system on the DISRPT benchmark. 
The system is trained on the \theory{} training set's contiguous DM and orphan token spans, with discontinuous spans split into two BIO-encoded connective instances. This means that discontinuous connectives (in accordance with PDTB's definitions, e.g.~\textit{if...then}) must be re-merged later, based on a closed list of discontinuous items attested in the training data. We also note that non-DM signals cannot be predicted using this system, since they often overlap. A DM-lexicon-based baseline is also provided for comparison in the next section.

\paragraph{Morphosyntax and Coreference} We use the AMALGUM pipeline \cite{gessler-etal-2020-amalgum}, which is designed to predict the same annotations present in GUM, including UD parses, entity annotations and coreference resolution. For testing we then use the same pre-processing scripts that feed the manual annotation for the \theory{} corpus described in Section \labelcref{sec:annotation}, except with predicted, rather than gold standard syntax trees and coreference, which can then be used to predict morphological, syntactic, semantic, and reference signals, and with no manual correction.

\paragraph{Association and Secondary Edges} Here we propose a new transformer-based text classifier, which receives two text spans known to be connected by a relation (based on the input primary tree), one of which contains a DM. The system predicts whether the relation between the spans is signaled by the DM, which is marked in the input by surrounding `**' characters. The spans are either the head EDUs of the relation (for intra-sentential relations) or the two sentences containing the head EDUs (for inter-sentential relations). We further embed the relation label, the distance between head EDUs and the direction of the relation in the input, and, for secondary edges, the relation label of any primary edge with the same source and target of the secondary relation, if available. The serialization is exemplified in \labelcref{ex:serialize}.

\ex. \textsc{antithesis} (\textsc{list}) \textit{left 1}: past studies have tended to avoid this task >> and have \textbf{**instead**} used samples of researchers\label{ex:serialize}

In \labelcref{ex:serialize}, the input suggests that a left-to-right secondary \textsc{antithesis} edge may exist between the given textual units, which are adjacent (direction and distance: \textit{left 1}), for which a primary \textsc{list} edge already exists, and which is marked by the orphan `instead'. Note that the DM `and', which also appears in the example (in fact, it is the DM for the primary \textsc{list} relation), is not targeted, as implied by the `**' notation, which singles out the word `instead'. 

In order to predict secondary edges at test time, the system also generates secondary edge prediction candidates for all primary edge paths attested in the predicted input tree, for all relation labels compatible with any predicted DM they contain, as well as relations between any adjacent pair of sentences, again provided that they contain a compatible DM. This compatibility is based on a DM-to-relation mapping obtained from the training data. Finally, the system ranks edges by binary classification probability and chooses the top possible relation to associate with each input DM (as predicted by DisCoDisCo). If the predicted relation is secondary then the DM is classified as an orphan, and the secondary edge is added to the graph. 

For the transformer embeddings we tested several options, trained on all true examples in the training set, enriched with an equal number of negative examples, and halting on dev set performance for early stopping. We compared base-sized versions of BERT, DeBERTa v3, XLNet and Electra, and chose Electra-base-discriminator as the highest performing model on the dev set.
\vspace{-6pt}

\subsection{Results}\label{sec:results}

Table \labelcref{tab:results-graph} gives results for \theory{} graph structures in seven scenarios. In the first row, we provide gold primary trees, syntax trees and connective positions, and the transformer model only predicts secondary edges and signal associations (\textit{RST=gold, NLP=gold}). This is an upper bound for the system performance, when no cascading errors from other components affect its accuracy.

In the other scenarios, we use AMALGUM tools for automatic syntax parsing and coreference resolution, and vary how primary trees and DMs are predicted. RST trees are either gold, or predicted using one of two RST parsers: the state of the art top-down parser DMRST (\citealt{liu-etal-2021-dmrst}) and, for comparison, the slightly less accurate, best bottom-up parser from \citet{guz-carenini-2020-coreference} (abbreviated G\&C). DMs are predicted either using a DM lexicon as a baseline (LX), or DisCoDisCo (DD) the SOTA system for DM detection. For the lexicon-based DM detection baseline we simply create a lexicon containing any string which is a DM in the training set more than $50$\% of the time and assume that it should always be predicted to be a DM, regardless of context (including multi-token DMs). 

In the bottom four scenarios in the table, only EDU segmentation and word-tokenization are given as inputs.\footnote{If these are not provided, numbers become very hard to compare due to segmentation conflicts; however we assume that both tasks can be performed automatically with high accuracy in production settings.} All tools are trained on the official GUM V9 training partition ($165$ documents), using the development partition for early stopping ($24$ documents) and the test set for the final scores ($24$ documents). As the table shows, secondary edge prediction is challenging even when gold RST trees and NLP pre-processing are given, especially for the \textsc{Full} metric. This is because correct prediction of a secondary edge requires the identification not only of a discourse relation (e.g.~that two parts of the text stand in \textsc{contrast}), but also that there is not already a primary edge corresponding to the relation, and that there is a sufficient trigger, such as an orphan DM or syntactic environment allowing for the secondary edge. Even with this information correctly recognized, the system must still choose the correct attachment points for the edge source and target in the hierarchical tree, as well as the edge direction and the label. Seen from this perspective, and considering the little training data available (fewer than 1K secondary edges), the \textsc{Span} score of $0.389$ is actually rather high, while the Relation score of $0.205$ is not much less than half the R score of a primary predicted parse $0.492$ (using DMRST, rows 4--5). This is despite the fact that primary parses gain score from easy wins, such as correct attachment of relative clauses and other explicit intrasentential relations -- secondary edges can be expected to be trickier cases.

\begin{table}[htb]
    \centering
    \small
\begin{tabular}{ll|llll|llll}
\toprule
& & \multicolumn{4}{c|}{Primary} & \multicolumn{4}{c}{Secondary} \\
\textbf{RST} & \textbf{NLP} & \textbf{S} & \textbf{N} & \textbf{R} & \textbf{F} & \textbf{S} & \textbf{N} & \textbf{R} & \textbf{F} \\
\midrule
gold & gold &      1.000 &      1.000 &      1.000 &      1.000 &      0.389 &      0.270 &      0.205 &      0.184 \\
gold & LX+AM &      1.000 &      1.000 &      1.000 &      1.000 &      0.210 &      0.142 &      0.113 &      0.091 \\
gold & DD+AM &      1.000 &      1.000 &      1.000 &      1.000 &      0.369 &      0.256 &      0.195 &      0.174 \\
DMRST & LX+AM &      0.620 &      0.545 &      0.492 &      0.482 &      0.055 &      0.044 &      0.027 &      0.022 \\
DMRST & DD+AM &      0.620 &      0.545 &      0.492 &      0.482 &      0.101 &      0.061 &      0.030 &      0.030 \\
G\&C-RST & LX+AM &      0.595 &      0.530 &      0.470 &      0.457 &      0.022 &      0.016 &      0.022 &      0.016 \\
G\&C-RST & DD+AM &      0.595 &      0.530 &      0.470 &      0.457 &      0.055 &      0.037 &      0.037 &      0.028 \\\bottomrule
\end{tabular}
    \caption{\theory{} graph metrics for our system with different inputs (LX=Lexicon-based connective detection; DD=DisCoDisCo connective detector; AM=UD trees and coreference from the AMALGUM parser; DMRST and G\&C are the top-down and bottom-up SOTA RST parsers cited above).}
    \label{tab:results-graph}
\end{table}

Turning to the impact of predictions by  baselines or previous SOTA tools, we see that automatically predicted NLP, including connective detection, does not produce substantial degradation in secondary edge predictions when DD is used, since the gold primary tree is still just as useful in determining whether a secondary edge is missing given existing primary ones, and connective detection is a relatively high performance task, often scoring over $90$\% for English \cite{braud-etal-2023-disrpt}. Using the LX baseline produces a very substantial degradation of almost half the score (second row). Predicted syntax trees could mainly impact prediction of syntactically motivated secondary edges (missing \textsc{attributions} from complement clauses, or \textsc{elaboration} relative clauses), yet these are not only rare, but also easy to predict correctly using a SOTA syntax parser. 

The situation in the last four rows is very different: Switching to predicted RST trees is catastrophic for secondary edge prediction, since, even if a relation missing from the primary tree is recognized, it could very well be an error in the primary parse: If the secondary edge detector correctly identifies and adds a real relation, the score will actually be impacted worse if that relation was a primary one in the gold data, since the detector then incurs both a precision and a recall error. Using a slightly less good parser does not matter as much as it does for the primary tree, but still degrades the Full metric (F). Switching to the LX baseline for DM detection is unsurprisingly catastrophic, especially when compounded with automatic primary RST parsing.

Moving on to the second part of the \theory{} graph prediction task, Table \labelcref{tab:results-signal-detection} shows performance on signal detection (identifying the signal types associated with each relation in the graph) and signal anchoring (also identifying the exact token span of each signal) broken down by major signal types, in the same scenarios. In each predicted scenario, the same scripts are used to identify the non-DM signal types for which automatic prediction is feasible, but the inputs are changed, e.g.~a syntactic relative clause signal is still predicted based on the syntax annotation, but in the predicted syntax setting, it uses automatic dependency parses, unlike in the gold data which we release with this paper.

\begin{table}[htb]
    \centering
    \resizebox{\textwidth}{!}{
\begin{tabular}{ll|llllllllll}
\toprule
 & & \multicolumn{10}{c}{\textsc{signal detection}} \\
 
 \midrule
\textbf{RST} & \textbf{NLP} & \textbf{all} & \textbf{dm} & \textbf{orphan} & \textbf{graph} & \textbf{morph} & \textbf{num} & \textbf{lex} & \textbf{sem} & \textbf{ref} & \textbf{syn} \\
 \midrule
gold & gold &      0.925 &      0.915 &      0.176 &      1.000 &      0.936 &      0.429 &      0.992 &      0.845 &      0.991 &      0.989 \\
gold & LX+AM &      0.756 &      0.724 &      0.080 &      0.886 &      0.957 & 0.0 &      0.870 &      0.686 &      0.802 &      0.961 \\
gold & DD+AM &      0.824 &      0.915 &      0.188 &      0.886 &      0.957 &      0.429 &      0.881 &      0.687 &      0.802 &      0.961 \\
DMRST & LX+AM &      0.450 &      0.351 &      0.030 &      0.416 &      0.500 & 0.0 &      0.435 &      0.271 &      0.137 &      0.838 \\
DMRST & DD+AM &      0.483 &      0.433 &      0.044 &      0.416 &      0.500 &      0.286 &      0.436 &      0.272 &      0.137 &      0.838 \\
G\&C-RST & LX+AM &      0.431 &      0.334 &      0.005 &      0.408 &      0.571 & 0.0 &      0.411 &      0.242 &      0.119 &      0.822 \\
G\&C-RST & DD+AM &      0.459 &      0.398 &      0.010 &      0.408 &      0.571 & 0.0 &      0.414 &      0.242 &      0.119 &      0.822 \\
\midrule

 & & \multicolumn{10}{c}{\textsc{signal anchoring}} \\

\midrule
gold & gold &      0.915 &      0.889 &      0.147 &      1.000 &      0.871 &      0.429 &      0.994 &      0.882 &      0.994 &      0.944 \\
gold & LX+AM &      0.555 &      0.679 &      0.055 &      0.972 &      0.900 & 0.0 &      0.837 &      0.459 &      0.537 &      0.898 \\
gold & DD+AM &      0.591 &      0.886 &      0.159 &      0.970 &      0.900 &      0.429 &      0.852 &      0.459 &      0.537 &      0.898 \\
DMRST & LX+AM &      0.298 &      0.331 &      0.017 &      0.567 &      0.386 & 0.0 &      0.416 &      0.138 &      0.088 &      0.786 \\
DMRST & DD+AM &      0.314 &      0.422 &      0.030 &      0.564 &      0.386 &      0.286 &      0.418 &      0.137 &      0.088 &      0.786 \\
G\&C-RST & LX+AM &      0.290 &      0.308 &      0.000 &      0.577 &      0.480 & 0.0 &      0.389 &      0.118 &      0.086 &      0.780 \\
G\&C-RST & DD+AM &      0.304 &      0.386 &      0.000 &      0.577 &      0.480 & 0.0 &      0.392 &      0.118 &      0.086 &      0.780 \\

\bottomrule
\end{tabular}
}
    \caption{\theory{} signal type detection and anchoring scores per signal category.}
    \label{tab:results-signal-detection}
    \vspace{-10pt}
\end{table}

As the table shows, here too DM results are quite good as long as the gold RST tree is provided and DD is used; the LX baseline produces substantially worse numbers for DMs/orphans and overall. With predicted primary RST trees, DM and orphan identities can again be swapped (if a primary/secondary relation pair are swapped in the prediction, what should be an orphan becomes a DM and vice versa), and in general, orphan prediction is challenging, since it only has a chance of being correct if the secondary edge was predicted correctly as well. We can also see that the penalty for switching to G\&C as the RST parser is fairly limited, but noticeable, mainly for DMs.

For non-DM signals too, predicted primary trees mean that the required relation for alignment may often not exist. This is especially clear for `easy' signal types, such as graphical ones, which include unambiguous punctuation and layout factors, such as bullet points marking a \textsc{list} relation -- if the structure of a \textsc{list} is predicted correctly, signal identification may be trivial, but an incorrect parse leads to a signal detection error as well.

Beyond these findings, we note that some signal types are challenging to get right even for gold trees, such as \textit{numerical} signals, which require matching numerical expressions to quantities of things mentioned, or \textit{morphological} ones, such as sequence of tenses. The latter signal type is predicted for any sequential temporal relation when units in sequence have succeeding tenses (past then present, present then future, etc.), but these morphological cues somewhere within the span of a sequence of events do not always indicate the sequence itself, as shown in example \labelcref{ex:bad-tense}, where a present tense direct speech predicate is uttered after a past tense narrative sentence, but the tense change is not actually a signal of the sequence.

\ex. $[$I \textbf{pulled} the bike to a halt (...)$]$ $[$``I \textbf{think} I've got a fairy stuck up my nose..''$]_{<sequence>}$ \label{ex:bad-tense}

NLP prediction quality, too, can matter considerably, even when gold RST trees are provided, for types such as \textit{reference} and \textit{semantics}, since automatic coreference resolution substantially underperforms the gold coreference information delivered in the gold NLP scenario. The most reliably predictable signal type is unsurprisingly \textit{syntactic}, for two reasons: (1) it depends on form-based NLP inputs for which reliable tools exist (syntactic dependency parsing), and (2) it is associated with some of the easiest relations to infer in the RST tree: Relative and other adnominal clauses, which both RST parsers usually parse correctly. 

In sum, these results demonstrate that while \theory{} parsing is a challenging task, it is not a hopeless one, especially in an era in which Computational Linguistics takes on increasingly complex tasks -- our system is a very rough proof of concept, and we are certain that better ones can be developed, even with current base-sized LMs, let alone much larger sequence-based LLMs. At the same time, it is clear that signal detection and secondary edge prediction is primarily feasible if we are confident we have the right primary tree: Without that tree, scores on the remaining tasks suffer from cascading errors very substantially. The same applies to a lesser extent to DM detection: with a better system for this subtask, scores on signals and secondary edges will rise, as made clear by the comparison between the LX baseline and DD.

\vspace{-3pt}

\section{Applications}\label{sec:applications}
\vspace{-1pt}

Although \theory{} parsing will require further research before we can expect to leverage reliable automatic analyses for practical applications, it is worth considering what information the formalism exposes and how it could be used in practice. Since \theory{} graphs can easily be reduced to primary unsignaled RST trees, it goes without saying that they provide the same benefits as those trees, e.g.~proposition extraction (=EDU segmentation); a built-in, recursive ability to extract the most prominent units in any document (or subspan) for extractive summarization, central discourse unit identification \cite{AtutxaEtAl2019}, topic segmentation \cite{xing-etal-2022-improving} or related tasks; and identification of specific relations of interest (e.g.~parsing all speeches of a public figure or political party and extracting all \textsc{concession} relations made by them for inspection), which can also be used for representation learning in downstream tasks \cite{huber-carenini-2022-towards,pu-etal-2023-incorporating}. Since relation spans and associated discourse markers are exposed by the graph, it is also possible to extract shallow discourse parses and use them to disambiguate connective senses or perform other tasks relying on shallow parsing, such as sentiment analysis or opinion mining. In fact, we are planning to leverage the in formation in \theory{} to generate training data compatible with current shallow discourse parsing frameworks as an additional resource in appropriate formats.

However, \theory{} graphs go beyond these original applications of discourse parsing to allow for additional, more fine-grained applications, for which we provide some examples here.\footnote{A reviewer has asked whether LLMs make such analyses redundant in practice. We do not believe so, for at least two reasons: 1. LLMs benefit from a variety of annotated data types for pre-training and instruction fine-tuning, and \theory{} data could be used to generate such supervision; and 2. identifying \theory{} relations with their associated signals is an end-task in itself, which can serve human analysts e.g.~in computational social science (analyzing political speech), quantitative and qualitative humanities research on rhetoric, and more, while offering evidence in support of the relations identified in each text.} In this section we would like to start by considering how much added value \theory{} brings to the core application of RST, i.e.~relation extraction, before considering some of its more novel applications and implications.

\paragraph{Relation Extraction compared to RST} At the most basic level, the addition of secondary edges allows analysts to represent multiple concurrent or tree-breaking relations without having to choose a single function label per unit in a text. Although secondary edges are comparatively rare, they are not evenly distributed across labels, and for some label types of interest, they may constitute over $10$\% of relation instances. Table \labelcref{tab:secrels} gives counts and proportions for secondary edges by relation label for labels that have >$5$ secondary instances in our data, in descending order of secondary edge percentage. The proportions give an idea of the extent of information an \theory{} parse gains (or a plain RST parse misses) out of the total possible relations recognized in our formalism. As the numbers show, primary trees alone miss substantial amounts of labels like \textsc{causal-result}, \textsc{explanation-justify}, and \textsc{adversative-concession}.

\begin{table}[h!tb]
\small
\centering
\begin{tabular}{l|rrrr}
\toprule
\textbf{Relation}  & \textbf{Primary} & \textbf{Secondary} & \textbf{Total} & \textbf{\% secondary} \\
\midrule
causal-result          & 437  & 72  & 509  & 14.10\% \\
explanation-justify    & 455  & 60  & 515  & 11.70\% \\
adversative-concession & 759  & 87  & 846  & 10.30\% \\
mode-manner            & 273  & 30  & 303  & 9.90\%  \\
adversative-antithesis & 375  & 35  & 410  & 8.50\%  \\
causal-cause           & 588  & 39  & 627  & 6.20\%  \\
elaboration-additional & 2,326 & 136 & 2,462 & 5.50\%  \\
joint-sequence         & 1,868 & 87  & 1,955 & 4.50\%  \\
contingency-condition  & 446  & 19  & 465  & 4.10\%  \\
elaboration-attribute  & 2,182 & 85  & 2,267 & 3.70\%  \\
context-circumstance   & 968  & 35  & 1,003 & 3.50\%  \\
explanation-evidence   & 729  & 26  & 755  & 3.40\%  \\
adversative-contrast   & 887  & 28  & 915  & 3.10\%  \\
joint-other            & 1,866 & 51  & 1,917 & 2.70\%  \\
joint-list             & 3,707 & 88  & 3,795 & 2.30\%  \\
joint-disjunction      & 305  & 6   & 311  & 1.90\%  \\
attribution-positive   & 1,335 & 24  & 1,359 & 1.80\%  \\
restatement-partial    & 370  & 5   & 375  & 1.30\%  \\
context-background     & 1,071 & 7   & 1,078 & 0.60\%  \\
\bottomrule
\end{tabular}
\caption{Proportions of secondary edges for relations with >5 secondary instances.} \label{tab:secrels} 
\vspace{-16pt}
\end{table}

\begin{figure}[h!tb]

  \includegraphics[clip,width=0.49\textwidth,trim = 0cm 0cm 0cm 0cm]{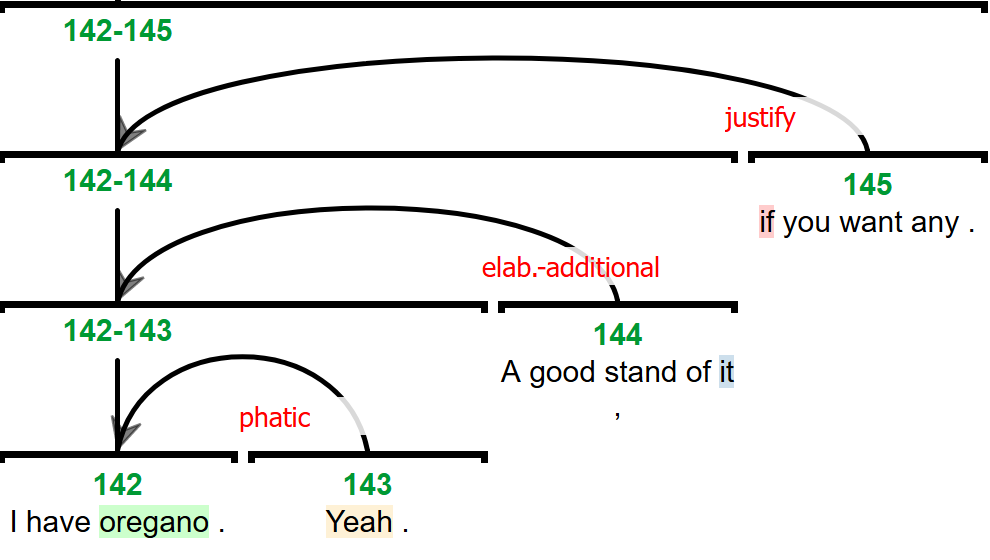}%
\centering

\caption{An \textsc{explanation-justify} relation marked by \textit{if} belongs to a special class of justifications which can be retrieved using signals, despite not having a dedicated relation label.}\label{fig:justify-if}
\vspace{-6pt}

\end{figure}

\paragraph{Signal-based Relation Subtypes} In addition to the word-sense disambiguation provided by associating DMs with relations (we can tell a temporal \textit{since} from a causal one, as in PDTB), signals can be used to identify subclasses of  relations which are more fine-grained than the two-level taxonomy used in our data. For example, RST-DT distinguishes a fine-grained relation \textsc{otherwise} (normally collapsed under the coarse \textsc{condition} class) which is not distinguished in GUM. However, using our anchored signals, it is easy to extract all relations marked by \textsc{otherwise} and obtain their exact satellite and nucleus scopes. It is also possible to define subtypes not found in any other datasets. For example, the non-conditional explanatory \textit{if} found in `I have oregano \textbf{if} you want any' characterizes a subtype of \textsc{explanation-justify} relation, which can be retrieved directly using the relation and discourse marker combination, as shown in Figure \labelcref{fig:justify-if} with the discourse marker in red.

Non-DM signals can also be used to extract subclasses of relation instances, such as identifying temporal relations signaled by explicit date or time expressions, \textsc{elaboration}s discussing meronyms of a nucleus entity, or \textsc{contrast} signaled via antonyms. In all of these cases, explicit signal annotations allow us to access sub-categories of relations, and even extract specific, open-class words, which expose more fine-grained semantic and pragmatic information. The potential of these possibilities is further enhanced by the presence of secondary relations, which can be queried concurrently to primary ones (e.g.~finding all \textsc{sequence} relations which are also an \textsc{elaboration}, or excluding cases that have a concurrent \textsc{contrast} relation, etc.).

\begin{figure}[h!tb]

  \includegraphics[clip,width=0.49\textwidth,trim = 0cm 0cm 0cm 0cm]{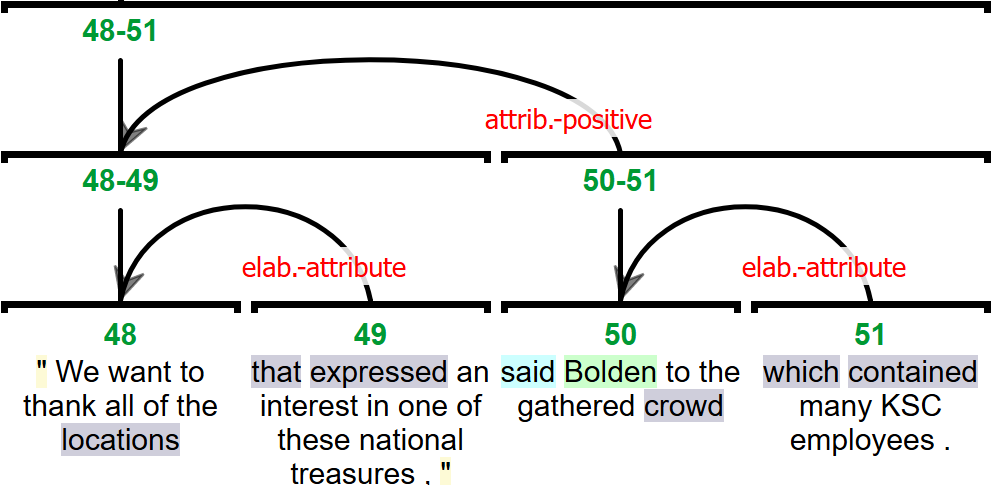}%
\centering

\caption{Attribution with anchored signals representing the attribution predicate in cyan and the attribution source in green.}\label{fig:attribution}

\vspace{-12pt}
\end{figure}

\paragraph{Attribution Scope, Source, Mode, and Polarity} Since attributions and their polarity are already identifiable using the relations \textsc{attribution-positive} and \textsc{attribution-negative} (e.g.~`officials did not say...'), and since the constituent tree expresses scope, RST data already exposes the span of positively or negatively attributed content. However, the addition of signals for \textit{attribution source} and  the \textit{indicative word} (or \textit{phrase}) instantiating the attribution predicate allow us to extract full information on the mode of attribution: Via a speech verb such as \textit{say} or cognitive predicate such as \textit{think}, or no predicate at all in `newspaper style' attribution giving just a quote and a name. The source of the attribution can correspond to a named or non-named entity. Figure \labelcref{fig:attribution} illustrates the information exposed by the formalism for a comprehensive extraction of attributions and their components: The \textit{attribution source} is marked in green and lexical predicate signal in cyan.

In the case of the multilayer GUM corpus, the existence of aligned lemmatization, entity recognition, entity linking (Wikification), and coreference resolution layers allow us to link attributions not only to an attested verb form (e.g.~\textit{said} in the figure, indicating the mode of attribution) and entity span (e.g.~\textit{Bolden}), but also to the predicate lemma (\textit{say}, substantially reducing the number of attribution mode predicate types) and the underlying entity identifier if available (e.g.~coreference cluster $3$, the cluster containing a mention \textit{NASA administrator Charles Bolden}, and also linked to the Wikipedia identifier \url{https://en.wikipedia.org/wiki/Charles_Bolden}).

\begin{figure}[h!tb]

\includegraphics[clip,width=0.7\textwidth,trim = 0cm 0cm 0cm 0cm]{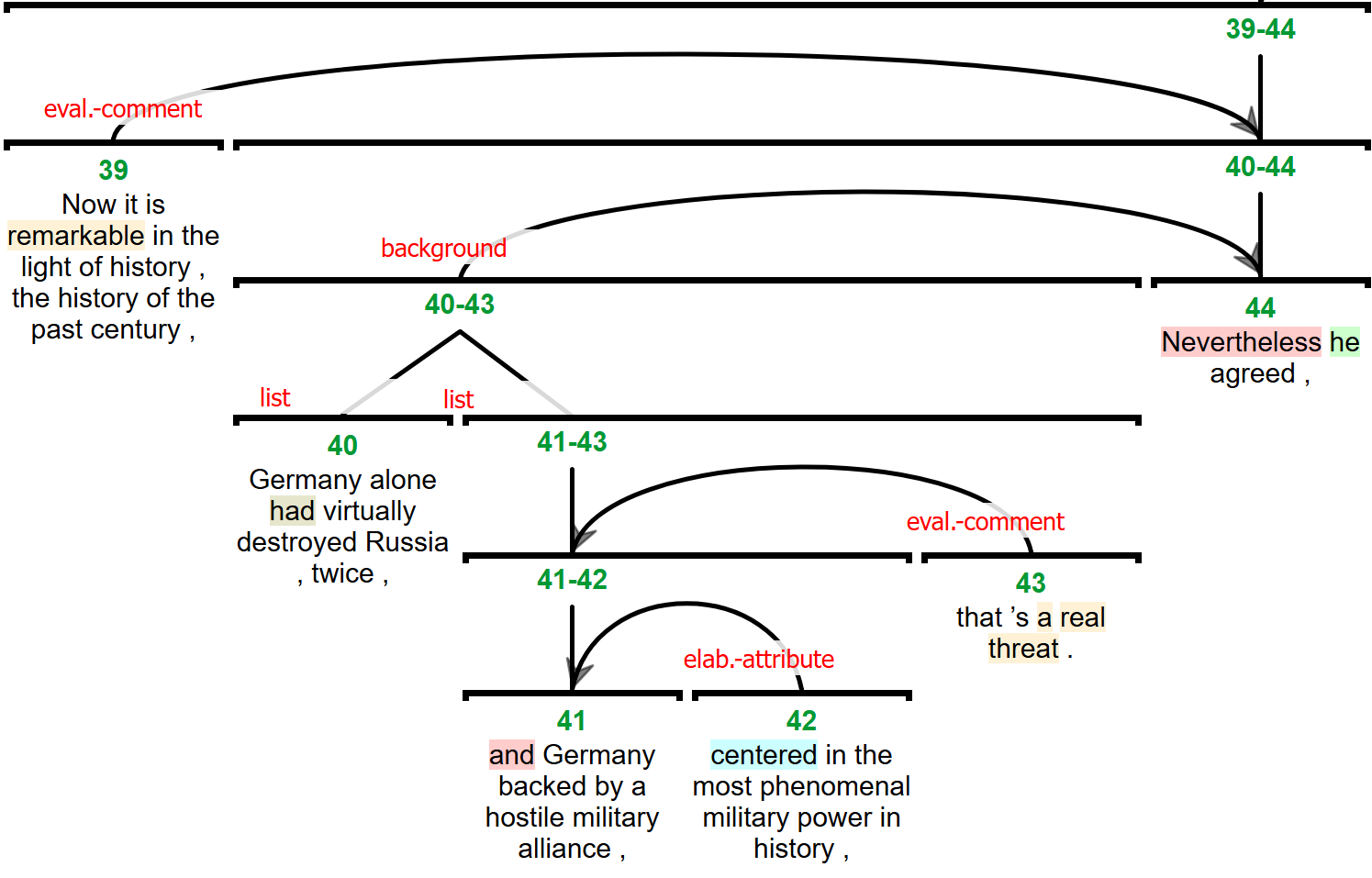}%
\centering
\caption{Two evaluations whose scope is marked by the tree, with the evaluation content signaled using indicative words and phrases: \textit{remarkable} and \textit{a real threat}.}\label{fig:evaluation}
\vspace{-8pt}
\end{figure}

\paragraph{Evaluation Content} Similarly to attributions, the relation \textsc{evaluation-comment} provides scope for what is being evaluated. However in a regular RST tree, it is not possible to know whether the evaluation is negative or positive (or neither), nor what evaluative terms were being used with respect to the content. The \theory{} graph improves on this whenever an \textit{indicative word} (or \textit{phrase}) is available, as illustrated in Figure \labelcref{fig:evaluation} in yellow.

The signaling annotations combined with the tree itself allow us to know that the entire span of units $40$--$44$ is being evaluated as \textit{remarkable}, while a nested evaluation from $43$ scoping over $41$--$42$ speaks of \textit{a real threat}. This information is substantially more detailed than what a basic RST tree can provide. Our data indicates that for $62.3$\% of \textsc{evaluation} relations, an associated indicative item is available, amounting to nearly $1,200$ tokens with over $200$ lemma types. The most common items are  in Table \labelcref{tab:eval-sigs}. Although top adjectives like `good' and intensifiers like `very' dominate the top of the table (amounting to nearly $14$\% of tokens), frequencies quickly drop to around $1$\%, reaching a frequency of $5$ at rank $56$, and single attestations (a.k.a.~hapax legomena) at rank 145. This shows the long tailed distribution of evaluative items, which are much less predictable or limited compared to DMs.

\begin{table}[h!tb]
\small
\centering
\begin{tabular}{rlrr}
\toprule
 \textbf{rank} & \textbf{lemma} & \textbf{frequency} & \textbf{\% signals} \\
 \midrule
1 & good & 90 & 0.076 \\
2 & very & 74 & 0.062 \\
3 & feel & 57 & 0.048 \\
4 & great & 30 & 0.025 \\
5 & important & 27 & 0.022 \\
6 & look & 27 & 0.022 \\
7 & bad & 23 & 0.019 \\
8 & seem & 23 & 0.019 \\
9 & mean & 22 & 0.018 \\
10 & beautiful & 20 & 0.016 \\
 & \multicolumn{3}{l}{...} \\
56 & fundamental & 5 & 0.004 \\
 & \multicolumn{3}{l}{...} \\
145 & inaccurate & 1 & 0.001 \\

\bottomrule
\end{tabular}
\caption{Frequency-ranked lemmas in lexical indicative signals for \textsc{evaluation}.} \label{tab:eval-sigs} 
\vspace{-15pt}
\end{table}

\paragraph{Reliability and Explainability} A major challenge for current neural NLP systems lies in reliability and explainability: When predicting structured outputs without a rationale, downstream applications and users have little way of knowing which predictions are likely to be correct or incorrect, what the rationale is for the prediction, and what we could do to filter out mistakes or improve systems. A complete automatic parse in \theory{} includes a built-in rationalization mechanism in the form of signals, which can be used for filtering (only use explicitly-signaled relations, or just ones signaled by a DM) and to better understand the predictions being outputted. Even though signal predictions can of course be wrong in themselves, especially explicit connective detection is now a fairly reliable task, and can be used by human analysts to better understand discourse parsing outputs, or as part of the input for downstream tasks which should be made aware of the strength and type of evidence for a system's predictions. By contrast, investigating cases of totally unsignaled relations in gold annotated data can help us to understand the limitations of our signaling annotation scheme, and try to address how human analysts arrive at an analysis in the absence of instances of anchorable, or even any signals of any kind.

\section{Conclusion}

In this article we presented \theory{}, a comprehensive theoretical framework representing discourse relations and structure, which expands on the existing Rhetorical Structure Theory, but incorporates insights from previous work in alternative frameworks, such as PDTB and SDRT. In particular, our proposal addresses weaknesses in RST, such as inability to handle tree-breaking and potentially multiple concurrent relations, as well as the failure to address the role of discourse relation marking devices, including, but not limited to, connectives. Going beyond PDTB's model, which is focused on morphosyntactically defined connectives and some additional highly constrained marker types, \theory{} adopts the view of the RST Signaling Corpus \cite{DasTaboada2017} by aspiring to a more exhaustive inventory of discourse relation signals, which we attach directly to relevant tokens in each text. The resulting representation retains advantages of RST, such as a strong commitment to recursive nuclearity and a hierarchical tree structure spanning entire documents, while enabling a more complete analysis covering previously disregarded relations, as well as the rationale for their identification in text-based terms.

Beyond the potential of \theory{} data to support more detailed theoretical studies of how discourse structure and meaning are constructed in natural language, we have also demonstrated some of the potential practical applications of \theory{}. These include not only classic uses of RST, such as searching for discourse relations (e.g.~finding \textsc{cause} and \textsc{result} in a parsed collection of texts, or \textsc{evidence} for a particular claim etc., now including tree-breaking cases), Central Discourse Unit detection, or extractive summarization, but also unique possibilities supported by the availability of signal annotations. The latter include tasks such as detailed \textsc{attribution} extraction, analysis of components of \textsc{evaluation} relations, and more. 

The introduction and definition of a parsing task for \theory{}, including revised evaluation metrics, a publicly available scorer, and a baseline implementation, provide a new and more comprehensive benchmark for discourse parsing, enriched by the presence of a more complete representation of the total relations available in each document. In particular, we hypothesize that the explicit presence of multiple concurrent relations in our data can shed more light on at least a subclass of parser errors in traditional RST parsing, in cases where parsers fail to predict the primary tree relations, but turn out to predict relations encoded as secondary ones, making errors that are not entirely wrong. We note that some recent initial work in this direction using double annotations in RST-DT seems to suggest that this hypothesis is correct \cite{liu-etal-2023-rsterrors}.

Additionally, we see potential for using data with the rich annotations present in \theory{} in probing and improving current LLMs, which can harness textual representations of the relations and signals exposed in our discourse parses for either pre-training tasks or for instruction fine-tuning. Zero or few-shot successes and failures in solving \theory{} tasks may also teach us about what specific LLMs do more or less well, and what levels of discourse awareness they possess. Conversely, we are optimistic that LLMs can be used to predict aspects of \theory{} parses, or eventually even complete parses, as some recent work on relation classification using sequence to sequence models has indicated \cite{anuranjana-2023-discoflan,chan-etal-2023-discoprompt}. Parsers built on top of LLM outputs may allow us to analyze larger datasets using increasingly accurate automatic parses, and to bootstrap data to tackle difficult, out-of-domain, or perhaps even multilingual scenarios in which manually annotated discourse parsing data is scarce.

Finally, we believe that the data and tools released with this article represent a substantial resource for research. The GUM corpus is now larger than the seminal RST-DT corpus for English \cite{CarlsonEtAl2001}, with $12$ written and spoken genres, showing the applicability of the framework to a broad range of text types. Since GUM continues to grow and cover new genres, we anticipate challenges but also opportunities in applying \theory{} to new kinds of data. We are also keenly aware of the limitation of richly annotated corpora primarily to English, and hope to be able to extend \theory{} to more languages in the future, with obvious first targets in the languages that already have RST treebanks which could be extended with \theory{} -- for example, the Georgetown Chinese Discourse Treebank \cite{peng-etal-2022-gcdt} follows the same RST annotation scheme as GUM, and many of the tools and scripts used for this article could be adapted to enrich it with relative ease (see Appendix \hyperref[app:gcdt]{B}). We hope that the release of the new \theory{} annotated GUM will encourage others to experiment with the framework and tools, and invite researchers working on discourse relations and representations to test the theory and provide feedback to evolve it further. For updates, annotation samples and discussion we also refer interested readers to the \theory{} website at \url{https://gucorpling.org/erst}.


\starttwocolumn
\bibliography{compling_style}

\appendix

\appendixsection{Relation Labels in GUM}\label{app:rels}

Table \labelcref{tab:gum-rels} gives the full list of relation labels in GUM. Note that \textsc{same-unit} is not a proper discourse relation, but rather a technical device used to connect multiple parts of a discontinuous EDU. For a fuller description of the labels and the most current GUM annotation guidelines, see \url{https://wiki.gucorpling.org/gum/rst/}. All definitions refer to the Reader (or hearer) as R, the Writer (or speaker) as W, a nucleus as N and a satellite as S. The nuclearity of the direction is either $\leftarrow$ (for satellite relations that only go left-to-right), $\rightarrow$ (the opposite), $\rightarrow\leftarrow$ (a satellite relation in either direction), or $\Lambda$ (multinuclear relation). Relation names all have the form <coarse-class>-<fine-grained>, i.e.~the first three relations in the table belong to the coarse class \textsc{adversative}.

\renewcommand\thetable{A.\arabic{table}}
\renewcommand\thefigure{B.\arabic{figure}}

\begin{table*}[bht]
\centering
\resizebox{\textwidth}{!}{%
\begin{tabular}{lll}
\toprule
\textbf{relation name}  & \textbf{nuclearity}  & \textbf{definition} \\
\midrule
\textsc{adversative-antithesis} & $\rightarrow\leftarrow$ & R is meant to prefer N as an alternative to S \\
\textsc{adversative-concession} & $\rightarrow\leftarrow$ & R is meant to look past an incompatibility of N with S \\
\textsc{adversative-contrast} & \hspace{0.6em}$\Lambda$ & W presents multiple Ns as incompatible, but of equal prominence \\
\textsc{attribution-negative} & $\rightarrow\leftarrow$ & S states that a potential source is NOT a source of the information in N \\
\textsc{attribution-positive} & $\rightarrow\leftarrow$ & S states a source for the information in N \\
\textsc{causal-cause} & $\rightarrow\leftarrow$ & S is the cause of N and N is more prominent) \\
\textsc{causal-result} & $\rightarrow\leftarrow$ & S is the result of N or: N is the cause of S, and N is more prominent) \\
\textsc{context-background} & $\rightarrow\leftarrow$ & S provides information to increase R's understanding of N \\
\textsc{context-circumstance} & $\rightarrow\leftarrow$ & S details circumstances often spatio-temporal) under which N applies \\
\textsc{contingency-condition} & $\rightarrow\leftarrow$ & N occurs depending on S \\
\textsc{elaboration-additional} & $\leftarrow$ & is used in all other cases, when S is an elaboration on N as a whole \\
\textsc{elaboration-attribute} & $\leftarrow$ & is used when S elaborates on a participant within N, rather than on the entire proposition in N \\
\textsc{evaluation-comment} & $\rightarrow\leftarrow$ & S provides an assessment of N by W R does not have to share this assessment) \\
\textsc{explanation-evidence} & $\rightarrow\leftarrow$ & S provides evidence which increases R's belief in N \\
\textsc{explanation-justify} & $\rightarrow\leftarrow$ & S increases R's acceptance of W's right to say N \\
\textsc{explanation-motivation} & $\rightarrow\leftarrow$ & S is meant to influence R's willingness to act according to N \\
\textsc{joint-disjunction} & \hspace{0.6em}$\Lambda$ & W presents multiple Ns which can be regarded as interchangeable alternatives \\
\textsc{joint-list} & \hspace{0.6em}$\Lambda$ & W presents multiple Ns in parallel which can be regarded as additive to one another  \\
\textsc{joint-other} & \hspace{0.6em}$\Lambda$ & any other collection of unlike discourse units of equal prominence at the same level of hierarchy \\
\textsc{joint-sequence} & \hspace{0.6em}$\Lambda$ & Multiple Ns form a temporally ordered sequence of events in order \\
\textsc{mode-manner} & $\rightarrow\leftarrow$ & S indicates the manner in which N happens \\
\textsc{mode-means} & $\rightarrow\leftarrow$ & S indicates the means by which N happens \\
\textsc{organization-heading} & $\rightarrow$ & explicit text organizing device such as a heading \\
\textsc{organization-phatic} & $\rightarrow\leftarrow$ & W holds the floor, without contributing propositional content \\
\textsc{organization-preparation} & $\rightarrow$ & covers all other forms of S units primarily used to signal an upcoming N \\
\textsc{purpose-attribute} & $\rightarrow\leftarrow$ & is used when S gives the purpose of a participant in N, rather than on the entire proposition in N \\
\textsc{purpose-goal} & $\rightarrow\leftarrow$ & the proposition in N as a whole is initiated or exists in order to realize S \\
\textsc{restatement-partial} & $\leftarrow$ & S partly realizes the same role and content as a previous N \\
\textsc{restatement-repetition} & \hspace{0.6em}$\Lambda$ & Multiple Ns realize the same role and content \\
\textsc{same-unit} & \hspace{0.6em}$\Lambda$ & indicates a discontinuous discourse unit this is not a discourse relation) \\
\textsc{topic-question} & $\rightarrow$ & N is the answer to the question posed by S \\
\textsc{topic-solutionhood} & $\rightarrow\leftarrow$ & N is a solution to a problem presented by S \\
\bottomrule
\end{tabular}
}
\caption{Relation Labels in the GUM Corpus.}
\label{tab:gum-rels}
\end{table*}

\appendixsection{GCDT Example}\label{app:gcdt}

\begin{CJK*}{UTF8}{gbsn}
\begin{figure*}[t]
\centering
\includegraphics[width=0.8\linewidth]{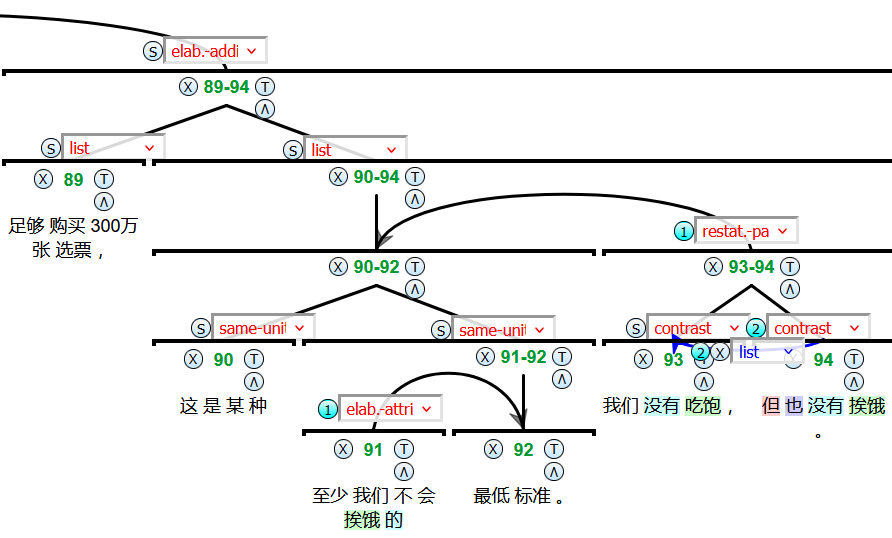}
\resizebox{0.8\textwidth}{!}{
\begin{tabular}{cll}
\toprule
EDU & Chinese & English translation \\
\midrule
89 & 足够 购买 300万 张 选票 ， & enough to buy 3 million votes \\
90 & 这 是 某 种  & this is some kind of \\
91 & 至少 我们 不 会 \underline{挨饿}\textsubscript{SEM\_REP}  
\underline{的}\textsubscript{SYN}  &  \underline{that}\textsubscript{SYN} at least we won't \underline{starve}\textsubscript{SEM\_REP} \\
92 &  最低 标准 。  &  minimum standard. \\
93 
&  \begin{tabular}[c]{@{}l@{}}
我们 \underline{没有}\textsubscript{SEM\_CHAIN} \underline{吃饱}\textsubscript{SEM\_ANT} ， 
\end{tabular}
&  \begin{tabular}[c]{@{}l@{}}
We \underline{are not}\textsubscript{SEM\_CHAIN} \underline{full}\textsubscript{SEM\_ANT}, 
\end{tabular}
\\
94 
&  \begin{tabular}[c]{@{}l@{}}
\underline{但}\textsubscript{DM} \underline{也}\textsubscript{ORPHAN} \underline{没有}\textsubscript{SEM\_CHAIN} \\
\underline{挨饿}\textsubscript{SEM\_REP, SEM\_ANT} 。\end{tabular} 
&   \begin{tabular}[c]{@{}l@{}} \underline{but}\textsubscript{DM} we \underline{are not}\textsubscript{SEM\_CHAIN} \\
\underline{starving}\textsubscript{SEM\_REP, SEM\_ANT} \underline{either}\textsubscript{ORPHAN}. 
\end{tabular}
\\
\bottomrule
\end{tabular}
}
\caption{GCDT example \textit{gcdt\_interview\_falkvinge}.}
\label{fig:gcdt}
\end{figure*}

Figure \labelcref{fig:gcdt} provides a sample eRST annotation in Mandarin Chinese using GCDT \citep{peng-etal-2022-gcdt}, along with translations of each EDU.
We can observe several discourse signals in this sample, 
for instance, a discourse marker~但~ \textit{d\`{a}n} ``but'', a semantic lexical chain 没有 \dots{} 没有\dots~\textit{m\'{e}i y\v{o}u... m\'{e}i y\v{o}u} ``not have \dots{} not have'',
an instance of semantic repetition~挨饿 \dots{} 挨饿~ \textit{\={a}i \`{e} \dots \={a}i \`{e}} ``starve \dots{} starving'',
and a semantic antonym ~吃饱 \dots{} 挨饿~\textit{ch\={\i} b\v{a}o \dots \={a}i \`{e}}
 ``full/eat enough \dots{} starving''. These function quite similarly to the equivalent examples from our English data, though the inventory and distribution of different signal types leaves much to study.

On top of these signals on primary relations, we also note the secondary edge between EDUs $93$ and $94$, which occurs in circumstances similar to English environments with multiple DMs, in this case, where an orphan 也 ~\textit{y\v{e}}~``also'' marks a secondary \texttt{list} relation. 
\clearpage
\end{CJK*}

\end{document}